\documentclass[10pt,journal,compsoc]{IEEEtran}
\ifCLASSOPTIONcompsoc
  \usepackage[nocompress]{cite}
\else
  \usepackage{cite}
\fi
\usepackage{amsmath,amssymb,amsfonts}
\usepackage{algorithmic}
\usepackage{algorithm}
\usepackage{bm}
\usepackage{graphicx}
\usepackage{textcomp}
\usepackage{xcolor}
\def\BibTeX{{\rm B\kern-.05em{\sc i\kern-.025em b}\kern-.08em
    T\kern-.1667em\lower.7ex\hbox{E}\kern-.125emX}}
\usepackage{url}
\usepackage{booktabs}
\usepackage{graphicx}
\usepackage{amsmath}
\usepackage{hyperref}
\hypersetup{colorlinks}
\usepackage{bm}
\usepackage{algorithm}
\usepackage{algorithmic}

\usepackage{amssymb}
\usepackage{amsthm}
\usepackage{enumerate}
\usepackage{multirow}
\usepackage{xcolor}
\newtheorem{theorem}{Theorem}

\newtheorem{definition}{Definition}

\author{Yi Xu, Weiran Shen, Jun Xu,~\IEEEmembership{Member,~IEEE,} , Xiao Zhang, \break and~Ji-Rong~Wen,~\IEEEmembership{Senior~Member,~IEEE}
\IEEEcompsocitemizethanks{
  \IEEEcompsocthanksitem Corresponding author: Xiao Zhang
  \IEEEcompsocthanksitem
  Y. Xu, W. Shen, J. Xu, X. Zhang, and J. Wen are with the Gaoling School of Artificial Intelligence, Renmin University of China, Beijing, China.\hfil\break
     E-mail:  \{yixu00, shenweiran, junxu, zhangx89, jrwen\}@ruc.edu.cn
}}

\title{IBCB: Efficient Inverse Batched Contextual Bandit for Behavioral Evolution History}

\begin{document}
\markboth{Journal of \LaTeX\ Class Files,~Vol.~14, No.~8, August~2015}{Xu \MakeLowercase{\textit{et al.}}: IBCB: Efficient Inverse Batched Contextual Bandit  for Behavioral Evolution History}

\IEEEtitleabstractindextext{%
\begin{abstract}
    Traditional imitation learning focuses on modeling the behavioral mechanisms of experts, which requires a large amount of interaction history generated by some fixed expert. However, in many streaming applications, such as streaming recommender systems, online decision-makers typically engage in online learning during the decision-making process, meaning that the interaction history generated by online decision-makers includes their behavioral evolution from \textit{novice expert} to \textit{experienced expert}. This poses a new challenge for existing imitation learning approaches that can only utilize data from experienced experts. To address this issue, this paper proposes an \textit{inverse batched contextual bandit} (IBCB) framework that can efficiently perform estimations of environment reward parameters and learned policy  based on the expert's behavioral evolution history. Specifically, IBCB formulates the inverse problem into a simple quadratic programming problem by utilizing the behavioral evolution history of the batched contextual bandit with inaccessible rewards, and it can be extended to fairness-aware expert limitation. We demonstrate that IBCB is a unified framework for both deterministic and randomized bandit policies. The experimental results indicate that IBCB outperforms several existing imitation learning algorithms on synthetic and real-world data and significantly reduces running time. Additionally, empirical analyses reveal that IBCB exhibits better imitation ability for fairness-aware experts, out-of-distribution generalization and is highly effective in learning the bandit policy from the interaction history of novice experts. The code is publicly available.
\end{abstract}
}

\maketitle
\IEEEpeerreviewmaketitle
\section{Introduction} \label{intro}
\IEEEPARstart{T}{raditional} imitation learning (IL) focuses on learning the decision-making policy of an experienced and fixed expert by using its historical behaviors. IL utilizes experts' demonstrations for policy fitting and aims to extract knowledge from experts' demonstrations to replicate their behaviors or environments. It can be divided into two main categories: Behavior Cloning (BC) \cite{Bain1995Framework} and Inverse Reinforcement Learning (IRL) \cite{Russell1998Learning}. BC approaches aim to learn a policy that directly maps states to actions, while IRL approaches focus on recovering the reward parameters from demonstrations to recover the experts' policy \cite{Zheng2022Imitation}. 

A major challenge in imitation learning (IL) is the substantial requirement of expert demonstration data \cite{Abbeel2004Apprenticeship, Hussein2017Imitation}. There is a common assumption that the experts are always experienced. 
However, in real-world streaming scenarios, the behavior policy of the expert constantly evolves over time, transitioning from a novice expert to an experienced one. Throughout this process, we accumulate a significant amount of expert's \emph{behavioral evolution history}.
 Taking streaming recommender systems as an example \cite{Zhang2021Counterfactual, Zhang2022Counteracting}, the agent in streaming recommendation needs to make a trade-off between exploitation and exploration during the recommendation process, and continuously and incrementally update its recommendation policy \cite{Chandramouli2011Streamrec, Chang2017Streaming, Jakomin2020Simultaneous}. 

In the behavioral evolution history, there exist a significant amount of contradictory data where an expert may take different actions when facing the same context at different time periods. This contradicts the assumption made by traditional imitation learning (IL) approaches regarding the consistency of expert behavior data. Meanwhile, recent works like \cite{wang2021fairness, gillen2018online} focus on maintaining fairness for bandit algorithms, which introduced a distribution-based random action selection, rather than simply selecting the best action, adding the difficulty to recover expert's parameters. Therefore, directly applying existing IL methods to recover the behavior policy from the expert's behavioral evolution history is not reasonable. For example, behavior cloning (BC) approaches are highly sensitive to distribution drift and contradictory data, and inconsistencies in the expert's evolution history can lead to a sharp decline in the final recovery performance.
Furthermore, some IL approaches rely on large-scale iterations of Metropolis-Hastings sampling to achieve accurate estimation in universal scenarios \cite{Ramachandran2007Bayesian, Choi2011MAP}. However, these approaches overlook the exploitation-exploration mechanism in expert interactions and demonstrate low sample efficiency when dealing with large-scale interaction data. While recent work by  \cite{Huyuk2022Inverse} focused on contextual bandits, it cannot be adapted to the batched bandit setting as it assumes the expert updates the policy in a fully-online manner. Additionally, existing inverse bandit approaches typically employ sampling techniques to estimate the likelihood, resulting in low training efficiency. 
There is an urgent need to develop inverse bandit methods specifically designed for the more general batched bandit setting and to devise novel techniques for recovering parameters tailored for bandit polices.

In this paper, we focus on the more general batched contextual bandit (BCB)  setting, and propose inverse batched contextual bandit (IBCB) to overcome several issues mentioned above. To learn from the evolving behaviors made by online decision-making experts with BCB policies, we design IBCB with a unified framework for both deterministic \cite{Han2020Sequential} and randomized \cite{Dimakopoulou2019Balanced} bandit policies. Under this unified framework, we take linear constraints with expectation relaxation in BCB's exploitation-exploration pairs and batched updating policy into consideration. We formulate the inverse problem of IBCB into a simple quadratic programming problem without notifying expert's behavior rewards. We take the assumption of BCB setting, rather than simply cloning behaviors with no assumption or blindly using large-scale samplings. Combining all these improvements together accelerates the train speed of IBCB and insures the robustness of IBCB for fairness-aware scenarios,  out-of-distribution and contradictory data. Due to that original BCB learns the reward parameter through online learning, IBCB can capture both expert's policy parameters and reward parameters at the same time. 

We summarize our major contributions:
(1) We define a new inverse bandit problem with behavioral evolution history in the BCB setting;
(2) We introduce a unified framework called IBCB, designed for both deterministic and randomized BCB policies, which can efficiently learn from the interaction history data of novice experts without requiring experts' feedbacks;
(3) IBCB is capable of simultaneously learning the parameters of the expert policy and the reward feedback, which outperforms various existing baseline IL approaches in scenarios such as fairness-aware experts, out-of-distribution data and contradictory data, with notable improvements in training speed.
\section{Related Works}

{\bf Contextual Bandits}
  have been extensively utilized for solving sequential decision-making problems in online learning \cite{Li2010Contextual, Lan2016Contextual, Yang2021Impact}. Recently, there has been increased research focus on a more general setting called batched contextual bandit (BCB) \cite{Han2020Sequential, Ren2020Dynamic, Gu2021Batched}, where actions within the same batch share fixed policy parameters. 
  In this regard, \cite{Han2020Sequential} proposed a UCB-like update policy for BCB, which is applicable to both random and adversarial contextual data.\\ 
{\bf Imitation Learning (IL)} is typically used to learn the expert's action selection policy or estimate the reward parameter from the expert's behavior \cite{Abbeel2004Apprenticeship, Hussein2017Imitation, Arora2021Survey, Zheng2022Imitation}. IL can be divided into two categories. The first category is Behavior Cloning (BC) \cite{Bain1995Framework}, which directly constructs a direct mapping from state to action 
\cite{Abbeel2010Autonomous, Osa2017Online, Liu2020Imitation}. The second category is the inverse reinforcement learning (IRL) method \cite{Russell1998Learning}, which attempts to recover environment's reward parameter from the expert's behavior, that is, imitating the reward parameters. Existing works mainly study IRL based on Bayesian \cite{Ramachandran2007Bayesian, Choi2011MAP, balakrishna2020policy}, maximum entropy (max-ent) \cite{Ziebart2008Maximum, Fu2018Learning, Qureshi2019Adversarial}, and generative \cite{Ho2016Generative, Chen2021Generative} approaches.
Traditional IL approaches mostly assume that the expert's behavior is optimal at every step \cite{Ziebart2008Maximum, Hussein2017Imitation, Brown2019Deep}, but ignore that early interactions are likely to be the evolutionary data of novice experts learned to become experienced experts.\\
{\bf Offline Reinforcement Learning (Offline RL)} has been widely used to improve the model's ability to generalize to out-of-distribution data, i.e., extrapolation error \cite{Prudencio2023Survey, Levine2020Offline}. There are two main research focuses in existing offline RL. One is to restrict the policy during training to avoid out-of-distribution generalization issues, which can be further subdivided into explicitly restricted policies \cite{Mnih:2015:DQN, Fujimoto2019Off, kumar2020conservative, kostrikov2021offline}and implicitly restricted policies \cite{Peng2019Advantage, Chen2020BAIL}. The other is to allow the learning model to adaptively estimate the uncertainty of the environment or data \cite{Kidambi2020MoReL, Yu2021COMBO}. 
Unlike Offline RL, which requires knowledge of the reward feedback, the proposed IBCB can learn about reward and expert policy's parameters from the history of expert behavioral evolution without requiring reward feedback information from the environment.

\section{Problem Formulation} \label{prob:form}
Let $[k]=\{1,2,...,k\}$, $\mathcal{S} \subseteq \mathbb{R}^d$ be the context space of dimension $d$, $[\bm{A};\bm{B}]=\left[\bm{A}^\top,\bm{B}^\top \right]^\top$, $\|\bm{x}\|_2$ denotes the $l_2$-norm of a vector $\bm{x}$. $\langle \cdot, \cdot \rangle$ denotes inner product of two vectors with same dimension.

\textbf{Batched Contextual Bandit (BCB) Setting.}  Following the setups in linear contextual bandit literature \cite{Dimakopoulou2019Balanced,Yang2021Impact,Li2010Contextual},
for any context $\bm s_{i} \in \mathcal{S} \subseteq \mathbb{R}^d$,
we assume that the expectation of the observed reward $R_{i}$ from environment is determined by unknown \emph{true reward parameters} $\bm \theta^* \in \mathbb{R}^d$:
$
     \mathbb{E} [  R_{i} ~|~  \bm s_{i} ] =   \langle \bm \theta^*, \bm s_{i} \rangle.
$ 
In this paper, our focus is on the generalized version of the traditional contextual bandit called the \textit{batched contextual bandit} (BCB) setting, which has gained considerable attention in the field of bandit theory and applications \cite{Han2020Sequential,Zhang2021Counterfactual}.
For an expert policy in BCB setting, we define expert's decision-making progress is partitioned into $N$ episodes, and in each episode, expert consists of two phases: (1) the \textit{online decision-making} chooses the action (i.e., candidate context, and also \textit{behavior}) for execution from each step's candidate context set following the updated and fixed expert policy $f$ for $B$ steps ($B$ is also called \textit{batch size}), and finally stores all context set \& action pairs and the observed rewards of the executed actions (contexts) into the data buffer $\mathcal{D}$ (\textit{expert's behavioral evolution history with rewards}); (2) the \textit{policy updating} approximates the optimal policy based on the received actions and rewards in $\mathcal{D}$. Note that for expert policy itself to evolve, expert can observe the reward of executed actions. Also, data buffer stores the indexes of the executed actions rather than actions themselves since each step's context set contains all actions, including the executed actions. This saves the storage space for practical applications. 

\begin{figure}
    \setlength{\belowcaptionskip}{-5pt}
    \centering
    \includegraphics[width=0.5\textwidth]{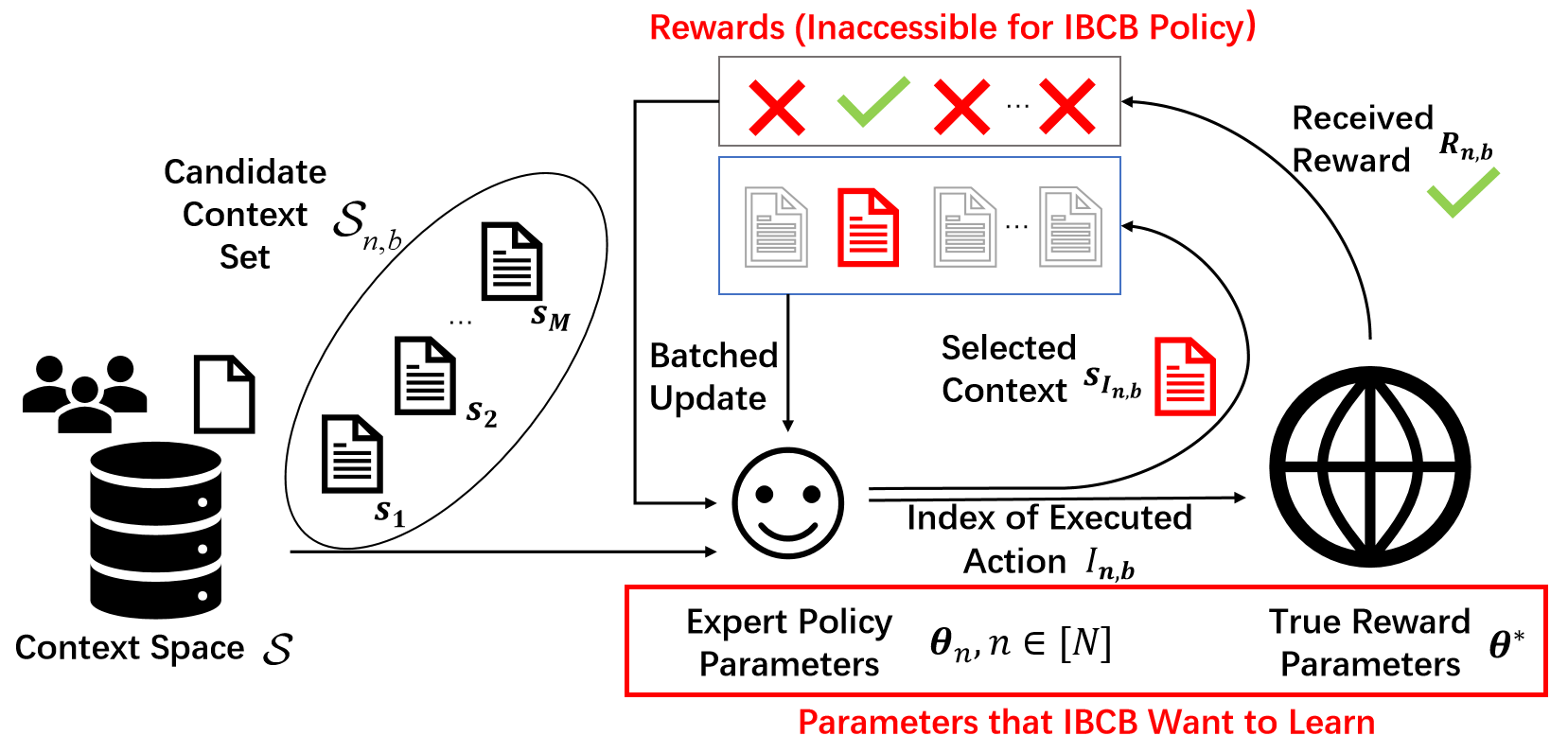}
    \caption{
    Inverse bandit problem with behavioral evolution history in BCB setting at step $b$ in $n$-th episode, where the rewards are inaccessible and our goal is to infer and estimate the policy parameters as well as the reward parameters. }
    \label{fig:evolution:estimation}

\end{figure}

\textbf{Inverse Bandit Problem.} Then, we introduce the formal definition of \textit{inverse bandit} problem with behavioral evolution history in BCB setting 
 as follows. Figure \ref{fig:evolution:estimation} shows the information that inverse batched contextual bandit~(IBCB) can observe, where reward of executed actions (\textit{behaviors}) is inaccessible. If expert do not update its parameter after each episode, this problem setting will degenerate to the traditional imitation learning problem.

\begin{definition}[Inverse Bandit Problem with Behavioral Evolution History in BCB Setting]
    Denote that $\bm s_{I}$, $I \in \mathcal{I}$, is a candidate context, $\mathcal{I}=\{I_j\}_{j \in [M]}$ is the action's index space containing \textit{M} action indexes, $\mathcal{S} \subseteq \mathbb{R}^d$ is the context space whose dimension is $d$, $\mathcal{S}_{n,b} = \{\bm{s}_I\}_{I \in \mathcal{I}} \subseteq \mathcal{S}$ is the candidate context set for expert at step $b$ in the $n$-th episode, $I_{n,b} \in \mathcal{I}$ is the index of the expert policy's executed action (behavior) at step $b$ in the $n$-th episode.
    Observe the behavioral evolution history generated by an expert with the format of $\mathcal{D}^{\mathrm{U}} = \bigcup_{n \in [N]} \mathcal{D}_{n}^{\mathrm{U}}$,
    where $\mathcal{D}_{n}^{\mathrm{U}} = \{ (\mathcal{S}_{n, b}, I_{n,b})\}_{b \in [B]}$, $\forall n \in [N]$, i.e., the reward feedbacks cannot be observed, $B$ is the batch size. We assume expert policy gets its parameters as $\{\bm{\theta}_{n}\}_{n \in [N]}$ through its online update method, and assume $\bm{\theta}_{N}$ is the closest estimation for reward parameter generated by expert policy in the evolution. We want to learn both expert policy parameter $\bm{\theta}_N$ and reward parameter $\bm{\theta}^*$ accurately through $\mathcal{D}^{\mathrm{U}}$. 
\end{definition}
\section{IBCB: The Proposed Approach}
In this section, we propose Inverse Batched Contextual Bandit approach named IBCB which can efficiently learn from expert's behavioral evolution history. 

\subsection{Policy Updating in BCB Setting} At the end of the $n$-th episode in the online learning period,
the context vectors (corresponding to the executed actions) and their rewards are observed, and are stored into a \emph{context matrix}
$\bm S_n \in \mathbb{R}^{B \times d}$ and a \emph{reward vector} $\bm R_n \in \mathbb{R}^B $, respectively.
In particular, each row index of $\bm S_n$ and $\bm R_n$ corresponds to the index of the step in which the context-reward pair was received. We introduce the updating process of the reward parameter vector $\bm \theta_{n} $ in BCB. We first concatenate the context and reward matrices from the previous episodes:
\begin{align*}
    &\bm L_{n}
    =  [\bm S_0; \bm S_1;
    \cdots; \bm S_n  ]\in \mathbb{R}^{nB \times d}, \\
    &\bm
    T_n = [ \bm R_0; \bm R_1;
    \cdots; \bm R_n ] \in \mathbb{R}^{nB}.
\end{align*}

Then, 
the \emph{updated parameter vector} $\bm \theta_{n+1}$ can be obtained by solving the following ridge regression: for $n = 0, 1, \ldots, N-1$,
\begin{equation}
\label{eq:batch:UCB:over_linear:share}
     \bm \theta_{n+1} = \operatornamewithlimits{arg\,min}_{\bm \theta \in \mathbb{R}^d}
      \left\| \bm L_n \bm \theta - \bm T_n  \right\|_2^2 +
      \lambda \| \bm \theta \|_2^2,
\end{equation}
where $\lambda >0$ is the regularization parameter.
The closed least squares solution of Eq.\eqref{eq:batch:UCB:over_linear:share} is used for estimating $\bm \theta_{n+1}$:
\begin{equation}\label{eq:PR:theta:origin:inc}
   \bm \theta_{n+1} =  \left( \bm  \Psi_{n+1} \right)^{-1} \bm b_{n+1},
\end{equation}
where $\bm \Phi_{n+1} = \bm \Phi_n + \bm S_{n}^\top  \bm S_{n} 
,\quad \bm b_{n+1} = \bm b_n + \bm S_{n}^\top  \bm R_n 
,\quad\bm  \Psi_{n+1} = \lambda \bm I_d + \bm \Phi_{n+1}.$

\begin{algorithm}[!htb]
\footnotesize
    \caption{Batched Policy Updating in the $n$-th episode in BCB Setting}   
    \label{alg:CBB:bandit:TS}
    \begin{algorithmic}[1]
    \REQUIRE Policy $f_{n}  $,
      data buffer $\mathcal{D}_{n}$,
        $\lambda \geq 0$,  
        $\alpha \geq 0$,
        $\bm \Phi_1  = \bm O_d$, $\bm b_1 =\bm 0,$ 
        batch size $B$, context space $\mathcal{S}$, action space $\mathcal{I}$
    \ENSURE  Updated policy $f_{n+1}$
          \STATE Store the selected context vectors and the observed rewards into
            $\bm S_n \in \mathbb{R}^{B \times d}$ and $\bm R_n \in \mathbb{R}^{B} $, respectively
            \STATE $\bm \Phi_{n+1} \leftarrow \bm \Phi_n + \bm S_{n}^{\top}  \bm S_{n},$ \quad
                $\bm b_{n+1} \leftarrow \bm b_n + \bm S_{n}^{\top}  \bm R_n$, \quad 
            $\bm \Psi_{n+1}  \leftarrow  \lambda \bm I_d + \bm \Phi_{n+1},$\quad
              $\bm \theta_{n+1} \leftarrow   ( \bm \Psi_{n+1}  )^{-1}  \bm b_{n+1} $
            \STATE  $// ~~ \texttt{Batched UCB Policy} ~~\texttt{(Step 5)}$
            \STATE  $f_{n+1} $ selects the context as
            $I_{n, b} \leftarrow \operatornamewithlimits{arg\,max}_{I \in \mathcal{I}}
            \langle \bm \theta_{n+1}, \bm s_I \rangle +
            \alpha [\bm s_I^\top \left(\bm \Psi_{n+1}\right)^{-1} \bm s_I]^{\frac{1}{2}}$, where $\bm s_I \in \mathcal{S}_{n, b}$
            \STATE  $// ~~ \texttt{Batched Thompson Sampling Policy} ~~\texttt{(Step 7 and 8)}$
            \STATE
            Draw $\tilde{\bm \theta}_{n+1}$ from $\mathcal{N}(\bm \theta_{n+1}, \alpha^2(\bm \Psi_{n+1})^{-1})$
            \STATE
            $f_{n+1} $ selects the context as
            $I_{n, b} \leftarrow \operatornamewithlimits{arg\,max}_{I \in \mathcal{I}}
            \langle \tilde{\bm \theta}_{n+1}, \bm s_I \rangle$,
            where $\bm s_I \in \mathcal{S}_{n, b}$
            \STATE \textbf{Return} $\bm \theta_{n+1} ,$~
                $  \left(\bm \Psi_{n+1}\right)^{-1} $
     \end{algorithmic}
\end{algorithm}
Finally, as shown in Algorithm~\ref{alg:CBB:bandit:TS}, the batched UCB policy updates the policy parameters and then selects the context $\bm s_{I_{n, b}}$ by applying the policy $f_{n+1}$ according to the following rule:
$I_{n, b} \leftarrow \operatornamewithlimits{arg\,max}_{I \in \mathcal{I}}
            \langle \bm \theta_{n+1}, \bm s_I \rangle +
            \alpha [\bm s_I^\top \left(\bm \Psi_{n+1}\right)^{-1} \bm s_I]^{\frac{1}{2}}$, 
where $\bm s_I \in \mathcal{S}_{n, b}$. 
On the other hand, the batched Thompson sampling policy is a  randomized policy that samples the policy parameters from a Gaussian distribution. However, as demonstrated in Theorem~\ref{thm:ICBCB:tsanducb}, we can prove that the batched Thompson sampling policy can also be expressed in the same regularized form as the UCB policy. Detailed proof can be found in Section~\ref{app:proof:thm:1}.
\begin{theorem}\label{thm:ICBCB:tsanducb}
For batched Thompson sampling policy in batched policy updating (i.e., step 7 and 8 in Algorithm~\ref{alg:CBB:bandit:TS}), we can obtain the reparameterized result of  $f_{n+1}$ as:
\begin{equation}\label{eq:reparam:SBTS:conclusion}
    I_{n, b} \leftarrow \operatornamewithlimits{arg\,max}_{I \in \mathcal{I}}
        \langle \bm \theta_{n+1}, \bm s_I \rangle + 
        \alpha [\bm s_I^\top \bm \Psi_{n+1}^{-1} \bm s_I]^{\frac{1}{2}} z
\end{equation}
where $z$ is a Gaussian random number drawn from $\mathcal{N}(0,1)$.
\label{thm:SBTS:reparam}
\end{theorem}

\subsubsection{Proof of Theorem~\ref{thm:SBTS:reparam}}
\label{app:proof:thm:1}
\begin{proof}
From Algorithm~\ref{alg:CBB:bandit:TS} we get the original form of SBTS's parameter:
\begin{equation}\label{eq:reparam:SBTS:condition:original}
    \tilde{\bm \theta}_{n+1} \sim \mathcal{N}(\bm \theta_{n+1}, \alpha^2(\bm \Psi_{n+1})^{-1})
\end{equation}
Since $\bm \Psi_{n+1}^{-1}$ is symmetric and positive-definite, so we can reparameterize $\tilde{\bm \theta}_{n+1}$ as follows:
\begin{equation}
    \tilde{\bm \theta}_{n+1} = \bm \theta_{n+1} + \alpha\bm B_{n+1}\bm z
\end{equation}
where $\bm B_{n+1}\bm B_{n+1}^\top=\bm \Psi_{n+1}^{-1}$, and $\bm z \sim \mathcal{N}(0,\bm{I}_{d})$.\\
Then we can rewrite the step 5 of Algorithm~\ref{alg:CBB:bandit:TS} as:
\begin{equation}\label{eq:reparam:SBTS:condition:after:reparam}
    I_{n, b} \leftarrow \operatornamewithlimits{arg\,max}_{I \in \mathcal{I}}
        \langle \bm \theta_{n+1}, \bm s_I \rangle + 
        \alpha \bm s_I^\top \bm B_{n+1}\bm z
\end{equation}

Note that for any matrix $\bm{A}$ with dimension of $r\times p$, $\bm{A}$ with dimension of $s\times q$, any random variable $x$ with dimension of $p \times 1$, $y$ with dimension of $q \times 1$ and any constant vector $b$ with dimension of $r \times 1$, $d$ with dimension of $s \times 1$, where $r,p,s,q$ is constant number, we have:
\begin{equation}
\label{eq:app:thm1:cov}
    \begin{aligned}
    Cov(\bm A_{r \times p}\bm x_{p \times 1}+\bm b_{r \times 1}, &\bm C_{s \times q}\bm y_{q \times 1}+\bm d_{s \times 1}) \\&= \bm A Cov(\bm x, \bm y) \bm C^\top
    \end{aligned}    
\end{equation}
where $Cov(\cdot,\cdot)$ means the covariance matrix.

Combining $\bm s_I^\top \bm B_{n+1}\bm z$ and \eqref{eq:app:thm1:cov} yields that:
 \begin{equation}
    \begin{aligned}
     Cov(\bm s_I^\top \bm B_{n+1}\bm z, \bm s_I^\top \bm B_{n+1}\bm z) & = \bm s_I^\top \bm B_{n+1}Cov(\bm z, \bm z) \bm B_{n+1}^\top\bm s_I \\
     & =\bm s_I^\top \bm B_{n+1} \bm B_{n+1}^\top \bm s_I \\
     & =\bm s_I^\top \bm \Psi_{n+1}^{-1} \bm s_I
    \end{aligned}
 \end{equation}
 
 And $\bm z \sim N(0, \bm{I}_{d})$, so we have:
 \begin{equation} \label{eq:app:thm1:cov:final}
     \bm s_I^\top \bm B_{n+1}\bm z \sim \mathcal{N}(0, \| \bm s_I^\top \bm \Psi_{n+1}^{-1} \bm s_I \|^2)
 \end{equation}
 Note that:
  \begin{equation} \label{eq:app:thm1:original}
      \left[\bm s_I^\top \bm \Psi_{n+1}^{-1} \bm s_I\right]^\frac{1}{2} z \sim \mathcal{N}(0, \| \bm s_I^\top \bm \Psi_{n+1}^{-1} \bm s_I \|^2)
 \end{equation}
 where $z \sim \mathcal{N}(0,1)$

So finally, after combining \eqref{eq:app:thm1:cov:final} and \eqref{eq:app:thm1:original}, we can rewritten \eqref{eq:reparam:SBTS:condition:after:reparam} as:
\begin{equation}\label{eq:reparam:SBTS:condition:final}
    I_{n, b} \leftarrow \operatornamewithlimits{arg\,max}_{I \in \mathcal{I}}
        \langle \bm \theta_{n+1}, \bm s_I \rangle + 
        \alpha [\bm s_I^\top \bm \Psi_{n+1}^{-1} \bm s_I]^{\frac{1}{2}} z
\end{equation}
\end{proof}

\subsection{IBCB's Formulation} 
We will illustrate the formulation using two actions as an example, which can be easily extended to accommodate multiple actions by adding additional constraints. 
Specifically, given $\mathcal{S}_{n+1, b} = \{ \bm s_1, \bm s_2\}$, $\mathcal{I}=\{1,2\}$, 
if action~1 
is executed at step $b$ in the $(n+1)$-th episode,
we have
\begin{align*}
            &\left\langle \left( \bm  \Psi_{n+1} \right)^{-1} \bm b_{n+1}, \bm s_1 \right\rangle +
            \alpha [\bm s_1^\top \left(\bm \Psi_{n+1}\right)^{-1} \bm s_1]^{\frac{1}{2}} \\
            &\geq
            \left\langle \left( \bm  \Psi_{n+1} \right)^{-1} \bm b_{n+1}, \bm s_2 \right\rangle +
            \alpha [\bm s_2^\top \left(\bm \Psi_{n+1}\right)^{-1} \bm s_2]^{\frac{1}{2}},
\end{align*}
which is equivalent to
\begin{equation}
\label{eq:inverse:UCB:condition:two}
\begin{aligned}
            &\left\langle \bm b_{n+1},  \left( \bm  \Psi_{n+1} \right)^{-1} (\bm s_1 - \bm s_2)  \right\rangle \\&\geq
            \alpha
            \left\{
                [\bm s_2^\top \left(\bm \Psi_{n+1}\right)^{-1} \bm s_2]^{\frac{1}{2}}
                -
                [\bm s_1^\top \left(\bm \Psi_{n+1}\right)^{-1} \bm s_1]^{\frac{1}{2}}
            \right\}.
\end{aligned}
\end{equation}
Setting $n = 1$ in Eq.\eqref{eq:inverse:UCB:condition:two},
from $\bm b_2 = \bm b_1 + \bm S_{1}^\top  \bm R_1= \bm S_{1}^\top  \bm R_1$ we get

\begin{equation}  
\label{eq:inverse:UCB:condition:two:yield}
\begin{aligned}
        &\left\langle      \bm R_1,   \bm S_{1}   \bm  \Psi_2^{-1} (\bm s_1 - \bm s_2)  \right\rangle \\&\geq
        \alpha
        \left\{
            [\bm s_2^\top  \bm \Psi_2^{-1} \bm s_2]^{\frac{1}{2}}
            -
            [\bm s_1^\top  \bm \Psi_2^{-1} \bm s_1]^{\frac{1}{2}}
        \right\},
\end{aligned}
\end{equation}

where $\bm s_1, \bm s_2 \in \mathcal{S}_{2, b}, \forall b \in [B].$ 
More generally, for all $ n \in [N-1]$,
given a candidate action set $\mathcal{S}_{n+1, b} = \{ \bm s_1, \bm s_2\}$,
if action 1 
is executed at step $b$ in the $(n+1)$-th episode (i.e., $\bm s_1$ is selected),
we can obtain that, for $\bm s_1, \bm s_2 \in \mathcal{S}_{2, b}, \forall b \in [B].$

\begin{equation}\label{eq:inverse:UCB:condition:two:yield:n}
    \begin{aligned}
        &\sum_{i \in [n]}
        \left\langle    \bm R_i,~
        \bm S_{i}   \bm  \Psi_{n+1}^{-1} (\bm s_1 - \bm s_2)  \right\rangle \\&\geq
        \alpha
        \left\{
            [\bm s_2^\top  \bm \Psi_{n+1}^{-1} \bm s_2]^{\frac{1}{2}}
            -
            [\bm s_1^\top  \bm \Psi_{n+1}^{-1} \bm s_1]^{\frac{1}{2}}
        \right\},
    \end{aligned}
\end{equation}

otherwise action 2 is selected for execution at step $b$ in the $(n+1)$-th episode.

Taking expectations of both sides of the inequality Eq.\eqref{eq:inverse:UCB:condition:two:yield:n},
we have

\begin{equation}\label{eq:inverse:UCB:condition:two:yield:n:exp}
    \begin{aligned}
        &\mathbb{E} \left[
        \sum_{i \in [n]}
        \left\langle    \bm R_i,~
        \bm S_{i}   \bm  \Psi_{n+1}^{-1} (\bm s_1 - \bm s_2)  \right\rangle
         \bigg| \bm S_1, \bm S_2, \ldots, \bm S_n
        \right]
        \\&\geq
        \alpha
        \left\{
            [\bm s_2^\top  \bm \Psi_{n+1}^{-1} \bm s_2]^{\frac{1}{2}}
            -
            [\bm s_1^\top  \bm \Psi_{n+1}^{-1} \bm s_1]^{\frac{1}{2}}
        \right\},
    \end{aligned}
\end{equation}

Combining linear contextual bandit's reward assumption in section \ref{prob:form}, Eq.\eqref{eq:inverse:UCB:condition:two:yield:n:exp} is equivalent to
\begin{equation}\label{eq:inverse:UCB:condition:two:yield:theta}
    \begin{aligned}
        &\sum_{i \in [n]}
        \left\langle    \bm \theta^*,~
        \bm S_i^\top  \bm S_{i}   \bm  \Psi_{n+1}^{-1} (\bm s_1 - \bm s_2)  \right\rangle \\&\geq
        \alpha
        \left\{
            [\bm s_2^\top  \bm \Psi_{n+1}^{-1} \bm s_2]^{\frac{1}{2}}
            -
            [\bm s_1^\top  \bm \Psi_{n+1}^{-1} \bm s_1]^{\frac{1}{2}}
        \right\}.
    \end{aligned}
\end{equation}

Under the constraints Eq.\eqref{eq:inverse:UCB:condition:two:yield:theta},
we can obtain the minimum-norm solution $\bm \theta^*$ by solving the following optimization problem with inequality constraints:
    \begin{align}
             ~~~~& \min \limits_{\bm \theta \in \mathbb{R}^d}~~
             \frac{1}{2}\| \bm \theta \|_2^2  \label{eq:inverse:opt:constriant}  \\
            &\text{s.t.}           ~
        \sum_{i \in [n]}
        \left\langle    \bm \theta,~
            \bm S_i^\top  \bm S_{i}
            \left[
                \bm \phi_{n}(\bm s_{\bar{I}_{n, b}}) -
                \bm \phi_{n}(\bm s_{I_{n, b}})
            \right]
        \right\rangle
        \\&\leq \alpha
        \left[
            H_n(\bm s_{I_{n, b}})
            -
            H_n(\bm s_{\bar{I}_{n, b}})
        \right]
        ,\ n\in [N], ~b \in [B], \nonumber
\end{align}
where  $\bm \phi_{n}(\bm s) = \bm \Psi_{n+1}^{-1} \bm s$, $H_n (\bm s) = \left[\bm s^\top \bm \phi_{n}(\bm s) \right]^{\frac{1}{2}}$. 
Eq.\eqref{eq:inverse:opt:constriant} is the final formulation of IBCB, which proposed a simple quadratic programming (QP) problem with large scale linear constraints. Several tools have been introduced to efficiently solve QP problems \cite{Gertz2003Object, Ferreau2014qpOASES, Stellato2020Osqp}. Among these approaches, OSQP \cite{Stellato2020Osqp} is the best to achieve almost linear complexity growth in problems' dimension for large scale low-accuracy Random QP problems, which can be applied to our IBCB. \\
When we assume that original expert algorithm is SBTS, we can simply apply a random parameter for $H_n(\bm s)$ as $H_n (\bm s) = \left[\bm s^\top \bm \phi_{n}(\bm s) \right]^{\frac{1}{2}}z$, where $z \sim \mathcal{N}(0, 1)$ by following Theorem~\ref{thm:SBTS:reparam} and clear $H_n (\bm s)$ to 0 by  following Eq.\eqref{eq:inverse:UCB:condition:two:yield:n:exp}'s expectation setting. This indicates that IBCB is a unified framework for both deterministic and randomized BCB.\\ 
\subsubsection{Algorithm Implementation}   
Existing inverse bandit algorithm \cite{Huyuk2022Inverse} estimates parameters using an EM-like algorithm, which iteratively maximizes the likelihood estimation. However, this algorithm suffers from low computational efficiency as it requires a large number of iterations and samples to achieve accurate estimation. Furthermore, the assumption of expert updates after each action under the fully-online setting leads to a rapid increase in computation time when sampling a large number of actions, making it unsuitable for the BCB setting. 
In contrast, the proposed IBCB capitalizes on the BCB assumption, which provides a unified framework for both deterministic and randomized bandit policies. By formulating the inverse bandit problem as a quadratic optimization problem with a significant number of linear constraints, IBCB offers a more efficient solution through leveraging the capabilities of the OSQP optimizer. Furthermore, since solving QP problems with full constraints costs high computation resources, and to better fulfill the updating method adopted in BCB assumption, we introduce the incremental implementation of IBCB algorithm. 

We provide pseudo code in the following part to describe the details of IBCB's incremental implementation in Algorithm~\ref{alg:incremental:IBCB:fulltext}.

\begin{algorithm}[!htb]
\footnotesize
    \caption{Incremental implementation of IBCB algorithm}   
    \label{alg:incremental:IBCB:fulltext}
    \begin{algorithmic}[1]
    \REQUIRE Initialize $\bm \theta_0 = \bm 0_d$, number of episodes $N$, batch size $B$,
      data buffer $\{\mathcal{D}_{n}\}_{n \in [N]}$,
        $\lambda = \lambda_0$,  
        $\alpha = \alpha_0$,
        $\bm \Phi_1  = \bm O_d$. 
    \FOR{$n = 1$ \TO $N$ }
    \STATE $// ~~ \texttt{Load previous data buffer}$
    \STATE Load last episode's selected context vectors $\bm S_{n} \in \mathbb{R}^{B \times d}$ from data buffer $\mathcal{D}_{n}$.

    \STATE Initialize current episode's constraint set $C_n = \emptyset$.
    \FOR{$b = 1$ \TO $B$}
    \STATE Load $b$-th step of $n$-th episode's selected item $I_{n,b}$, other candidate items $\bar{I}_{n,b}$ and all candidate context vectors $\bm s$.\\
    \STATE $\bm \Phi_{n+1} \leftarrow \bm \Phi_n + \bm S_{n}^{\top}  \bm S_{n},$ 
    \STATE $\bm \Psi_{n+1}  \leftarrow  \lambda \bm I_d + \bm \Phi_{n+1}$.
    \STATE  $// ~~ \texttt{Construct constraints}$
    \STATE Construct current step's constraint element $c$ as: \\
    $c \leftarrow  \{\sum_{i \in [n]}
        \left\langle    \bm \theta,~
            \bm S_i^\top  \bm S_{i}
            \left[
                \bm \phi_{n}(\bm s_{\bar{I}_{n, b}}) -
                \bm \phi_{n}(\bm s_{I_{n, b}})
            \right]
        \right\rangle
        \leq \alpha
        \left[
            H_n(\bm s_{I_{n, b}})
            -
            H_n(\bm s_{\bar{I}_{n, b}})
        \right]\}$, \\where  $\bm \phi_{n}(\bm s) = \bm \Psi_{n+1}^{-1} \bm s$, $H_n (\bm s) = \left[\bm s^\top \bm \phi_{n}(\bm s) \right]^{\frac{1}{2}}$.
        
    \STATE Update: $C_n \leftarrow C_n \cup c$.
    \ENDFOR
    \STATE  $// ~~ \texttt{Solve the QP problem}$
    \STATE Solve the following optimization problem with current episode's constraint set $C_n$ through OSQP optimizer:\\
    $\min \limits_{\bm \theta \in \mathbb{R}^d}  \frac{1}{2}\| \bm \theta  - \bm \theta_{n-1} \|_2^2 \quad \text{s.t.}\   C_n$
    \STATE $// ~~ \texttt{Incrementally update}$
    \STATE Use solved parameter $\bm \theta$ to update: $\bm \theta_{n} = \bm \theta$.
    \ENDFOR
    \STATE \textbf{Return} Learned expert policy parameter $\bm \theta_{N}$ (reward parameter $\bm \theta^{*}$)
     \end{algorithmic}
\end{algorithm}

\subsubsection{Analysis of time complexity of IBCB}
After using incremental implementation, overall time complexity of IBCB solution is reduced from $O((NBM)^\beta d)$ (non-incremental IBCB algorithm) to $O(N (BM)^\beta d)$ (incremental IBCB algorithm), where $N$ is the total number of episodes, $B$ is the total number of steps in each episode, $M$ is the size of the action space, and $d$ is the context space dimension.  $\beta$ is the adaptive complexity parameter corresponding to the QP problem of different constraints for OSQP. According to the experimental results and proof of OSQP, $\beta \in \left[1,2\right]$. When the quantity of constraints is fixed, $\beta$ will adjust adaptively. When there are moderate constraints ($\leq 10^6$ constraints), $\beta$ will tend to 1. Analysis can be found in the following part.

We states that the training time complexity of IBCB is $O((NBM)^\beta d)$ (non-incremental IBCB) to $O(N (BM)^\beta d)$ (incremental IBCB), where $N$ is the total number of episodes, $B$ is the total number of steps in each episode, $M$ is the size of the action space, and $d$ is the context space dimension. $\beta$ is the adaptive complexity parameter corresponding to the QP problem of different constraints for OSQP. According to experimental results and the proof of OSQP [4], $\beta \in \left[1,2\right]$. When the number of constraints is fixed, $\beta$ adjusts adaptively and tends to 1 when there are moderate constraints ($\leq 10^7$ constraints).\\ 
\textit{Proof.} For any $n \in N$, i.e., in the $n$-th episode,\\
1. Solving matrix $\phi_{n+1}$ in Algorithm 2 in Appendix A.1 requires matrix multiplication, with a time complexity of $O(d\*B\*d) = O(Bd^2)$.\\
2. Constructing $B*(M-1)$ constraints for each step in an episode with $M$ actions requires a time complexity of $O(BMd^3)$.\\
3. Using the current $B*(M-1)$ constraints to solve with OSQP requires a time complexity of $O((BM)^\beta d)$, where $\beta$ is the adaptive complexity parameter, $\beta \in \left[1,2\right]$, close to 1 when constraints $\leq 10^7$ Therefore, in the incremental update training of the IBCB algorithm, the time complexity of one episode is $O(Bd^2 + BMd^3 + (BM)^\beta d)$. Considering that $d$ is usually less than 100 in practice, the basic unit time for matrix multiplication and inversion is very low. Therefore, the time complexity of one episode is $O(BM+(BM)^\beta d) = O((BM)^\beta d)$. So, in the setting of a total of $N$ episodes, the training time complexity of the incremental IBCB is $O(N(BM)^\beta d)$, where $\beta$ is the adaptive complexity parameter, $\beta \in \left[1,2\right]$, close to 1 when constraints $\leq 10^7$. \\
4. If the incremental learning algorithm is not used to implement IBCB, i.e. non-incremental IBCB, OSQP needs to solve $NB*(M-1)$ constraints simultaneously, so the total time complexity of the non-incremental implementation is $O((NBM)^\beta d)$.
\subsubsection{Extension to Fairness-Aware Expert Limitation} 
According to \cite{wang2021fairness}, a fairness-aware expert selects actions based on a nonlinear transformation of their selection probabilities, such as using Softmax, to ensure fairness by not only execute the best action. However, this can significantly reduce the average reward. To address this, we've enhanced the \textbf{top-$k$} contextual bandit recommendation algorithm from \cite{mansoury2022exposure} with a compromise policy: choosing the final action from the \textbf{top-$k$} based on normalized probabilities, which balances reward and fairness. 
Our IBCB can be extended to imitate the above fairness-aware expert.  
Specifically, during each training episode, IBCB first trains for one round to obtain the initial reward parameters $\bm \theta_\tau$ in current episode $n$. IBCB then uses $\bm \theta_\tau$ to clean up the inequality constraints in Eq.~\eqref{eq:inverse:opt:constriant} as a pre-process method, removing constraints that do not meet the criteria since expert may not select the best action
, and subsequently retrains for one episode based on the remaining constraints. 

Specifically, considering that IBCB has a corresponding incremental learning framework, combined with the update method of the BCB for each episode, we have made a new modification to IBCB after adding fairness. Details can be found in the following pseudo code in Algorithm~\ref{alg:IBCB:fair:fulltext}.

\begin{algorithm}[!t]
\footnotesize
    \caption{IBCB's adaptation to fairness-aware expert limitation}   
    \label{alg:IBCB:fair:fulltext}
    \begin{algorithmic}[1]
    \REQUIRE Initialize $\bm \theta_0 = \bm 0_d$, number of episodes $N$, batch size $B$,
      data buffer $\{\mathcal{D}_{n}\}_{n \in [N]}$,
        $\lambda = \lambda_0$,  
        $\alpha = \alpha_0$,
        $\bm \Phi_1  = \bm O_d$. 
    \FOR{$n = 1$ \TO $N$ }
    \STATE $// ~~ \texttt{Load previous data buffer}$
    \STATE Load last episode's selected context vectors $\bm S_{n} \in \mathbb{R}^{B \times d}$ from data buffer $\mathcal{D}_{n}$.

    \STATE Initialize current episode's constraint set $C_n = \emptyset$.
    \FOR{$b = 1$ \TO $B$}
    \STATE Load $b$-th step of $n$-th episode's selected item $I_{n,b}$, other candidate items $\bar{I}_{n,b}$ and all candidate context vectors $\bm s$.\\
    \STATE $\bm \Phi_{n+1} \leftarrow \bm \Phi_n + \bm S_{n}^{\top}  \bm S_{n},$ 
    \STATE $\bm \Psi_{n+1}  \leftarrow  \lambda \bm I_d + \bm \Phi_{n+1}$.
    \STATE  $// ~~ \texttt{Construct constraints}$
    \STATE Construct current step's constraint element $c$ as: \\
    $c \leftarrow  \{\sum_{i \in [n]}
        \left\langle    \bm \theta,~
            \bm S_i^\top  \bm S_{i}
            \left[
                \bm \phi_{n}(\bm s_{\bar{I}_{n, b}}) -
                \bm \phi_{n}(\bm s_{I_{n, b}})
            \right]
        \right\rangle
        \leq \alpha
        \left[
            H_n(\bm s_{I_{n, b}})
            -
            H_n(\bm s_{\bar{I}_{n, b}})
        \right]\}$, \\where  $\bm \phi_{n}(\bm s) = \bm \Psi_{n+1}^{-1} \bm s$, $H_n (\bm s) = \left[\bm s^\top \bm \phi_{n}(\bm s) \right]^{\frac{1}{2}}$.
        
    \STATE Update: $C_n \leftarrow C_n \cup c$.
    \ENDFOR
    \STATE  $// ~~ \texttt{Solve the QP problem}$
    \STATE Solve the following optimization problem with current episode's constraint set $C_n$ through OSQP optimizer:\\
    $\min \limits_{\bm \theta \in \mathbb{R}^d}  \frac{1}{2}\| \bm \theta  - \bm \theta_{n-1} \|_2^2 \quad \text{s.t.}\   C_n$
    \STATE $// ~~ \texttt{Incrementally update}$
    \STATE Use solved parameter $\bm \theta$ as this episode's initial reward parameter $\bm \theta_\tau$: $\bm \theta_{\tau} = \bm \theta$.
    \STATE $// ~~ \texttt{Use initial reward parameter -}$
    \STATE $// ~~ \texttt{- to clean up unsatisfied constraints}$
    \STATE Initialize current expisode's unsatisfied constraint set $C_{n-} = \emptyset$.
    \FOR{$b = 1$ \TO $B$}
    \IF{$ \sum_{i \in [n]}
        \left\langle    \bm \theta_{\tau},~
            \bm S_i^\top  \bm S_{i}
            \left[
                \bm \phi_{n}(\bm s_{\bar{I}_{n, b}}) -
                \bm \phi_{n}(\bm s_{I_{n, b}})
            \right]
        \right\rangle
        > \alpha
        \left[
            H_n(\bm s_{I_{n, b}})
            -
            H_n(\bm s_{\bar{I}_{n, b}})
        \right]$}
    \STATE Construct current step's unsatisfied constraint element $c_{-}$ as: \\
    $c_{-} \leftarrow \{\sum_{i \in [n]}
        \left\langle    \bm \theta_{\tau},~
            \bm S_i^\top  \bm S_{i}
            \left[
                \bm \phi_{n}(\bm s_{\bar{I}_{n, b}}) -
                \bm \phi_{n}(\bm s_{I_{n, b}})
            \right]
        \right\rangle
        > \alpha
        \left[
            H_n(\bm s_{I_{n, b}})
            -
            H_n(\bm s_{\bar{I}_{n, b}})
        \right]\}$
    \STATE Update unsatisfied constraint set: $C_{n-} \leftarrow C_{n-} \cup c_{-}$.
    \ENDIF
    \ENDFOR
    \STATE Initialize current episode's satisfied constraint set $C_{n}^{'}$ by removing unsatisfied constraints: $C_{n}^{'} \leftarrow C_n \backslash C_{n-}$.
    \STATE  $// ~~ \texttt{Solve QP problem -}$
    \STATE  $// ~~ \texttt{- with satisfied constraints}$
    \STATE Solve the following optimization problem with current episode's satisfied constraint set $C_{n}^{'}$ through OSQP optimizer:\\
    $\min \limits_{\bm \theta^{'} \in \mathbb{R}^d}  \frac{1}{2}\| \bm \theta^{'}  - \bm \theta_{n-1} \|_2^2 \quad \text{s.t.}\   C_{n}^{'}$
    \STATE $// ~~ \texttt{Incrementally update}$
    \STATE Use solved parameter $\bm \theta^{'}$ to update: $\bm \theta_{n} = \bm \theta^{'}$.
    \ENDFOR
    \STATE \textbf{Return} Learned expert policy parameter $\bm \theta_{N}$ (reward parameter $\bm \theta^{*}$)
     \end{algorithmic}
\end{algorithm}


\section{Experiments}
\subsection{Baselines}
IBCB was compared with several existing original expert algorithms and 
imitation learning (IL) algorithms, including:
\subsubsection{Original Expert Algorithms with Rewards (SBUCB \& SBTS)}
We used SBUCB and SBTS for original expert algorithms. These algorithms' performances are the oracle limit above all of other IL algorithms and IBCB since they have access to the rewards. SBUCB is a batched version of LinUCB \cite{Li2010Contextual}, which updates the policy after receiving a batch of feedback data \cite{Han2020Sequential}. SBTS is a batched version of Linear Thompson Sampling (LinTS) \cite{Dimakopoulou2019Balanced}.
\subsubsection{Behavior Cloning (BC) Algorithms} 
 We used LinearSVC \cite{Pedregosa2011Scikit} model (namely BaseSVC) for BC baseline, since the we assumed reward enviornment is linear in Section \ref{prob:form}.
\subsubsection{Inverse Reinforcement Learning (IRL) Algorithms} 
We used Bayesian Inversive Reinforcement Learning (B-IRL) \cite{Ramachandran2007Bayesian}, and current state-of-the-art Bayesian Inversive Contextual Bandits (B-ICB) \cite{Huyuk2022Inverse} to serve as IRL baselines.

\subsection{Datasets}

We designed synthetic and real-world dataset for experiments. For both of them, we split the whole dataset into two subsets: the Online Learning (the OL-data) phase, and the Batch Testing (the BT-data) phase. Note that expert only update its policy parameters in OL-data, and do not update parameter in BT-data phase so we could design experiments on estimating expert policy's parameters on BT-data.

\subsubsection{Synthetic Dataset}
\label{exp:dataset:syn}
 We set the synthetic dataset's candidate context $\bm s_{I}$ with context dimension $d=10$ as a single context (behavior) for any step of an episode. At step $b$ of $n$-th episode in OL-data phase, candidate  context set $\mathcal{S}_{n,b} \subseteq {\mathbb{R}^d}$ was drawn from a Gaussian distribution $\mathcal{N}(\mu_s\boldsymbol{1}_d, \sigma_s^2\boldsymbol{I}_d)$, where the means of candidate contexts were $\mu_s\in\left[1:-0.4:-2.6\right]$, and the standard deviation was $\sigma_s^2={0.05}^2$. To simulate out-of-distribution(OOD) scenario, the mean of the first candidate context in the BT-data phase was set to $1.4$. We set the observed reward of a given candidate context $\bm s_{I_{n,b}} \in \mathcal{S}_{n,b}$ at step $b$ in $n$-th episode as a sigmoid function $\mathrm{sigmoid}(\left\langle \bm{w}_{\mathrm{reward}}, \boldsymbol{s}_{I_{n,b}}\right\rangle)$, where each element of $\bm{w}_{\mathrm{reward}} \in \mathbb{R}^d$ was sampled from a Gaussian distribution as $\mathcal{N}(0.1, {0.01}^2)$. To test the robustness of the model and simulate random selections of actions made by users, we also added a zero-mean Gaussian noise $\mathrm{std} \sim \mathcal{N}(0,\epsilon^2), \epsilon \in \{0.03, 0.07,0.10\}$ in the reward, so final reward was set to be $\mathrm{sigmoid}(\left\langle \bm{w}_{\mathrm{reward}}, \boldsymbol{s}_{I_{n,b}}\right\rangle) + \mathrm{std}$. We also made adjustments for the ablation study period. Contradictory data was introduced as \textit{duplicated data}: we cloned the first \textit{k} episodes' candidate context sets for the second \textit{k} episode's candidate context sets, and design \textit{k} with the name of \textit{duplicated quantity} $\mathrm{dup} \in \left[0:1:9\right]$. Total number of conflict executed actions (\textit{e.g. expert select action 1 at step $b$ of $n$-th episode, but select action 2 at step $b$ of $(n+k)$-th episode.}) of the contradictory data was ascending when we increased $\mathrm{dup}$. We also controlled the training rate for all baselines and IBCB: we set the end episode of the exposed expert's behavioral evolution history logs with constraint end rate $\mathrm{ce\_rate} \in \left[0,1\right]$ ($e.g.$ if $\mathrm{ce\_rate}=0.5$. we used the first $\left\lfloor0.5N\right\rfloor$ episodes for baselines and IBCB training).

 In a nutshell, we generated the synthetic data for both OL-data and BT-data as follows: numbers of episodes $N=20$, batch size $B=1000$, number of candidate contexts (actions) is $M=10$ at any single step of an episode.

\subsubsection{Real-World Dataset}
\label{exp:dataset:real}
We used MovieLens 100K \footnote{\url{https://grouplens.org/datasets/movielens/100k/}} (ML-100K) dataset as our real-world experiment dataset. 
, which collected 5-star movie ratings from a movie recommendation service, including 943 unique users and 1682 unique movies (items). ML-100K dataset also provided various side information of both users and items. We split ML-100K dataset into OL-Data and BT-data according to the timestamps, where first half was treated as OL-data and the last half was BT-data. For reward setting, we treated the 4-5 star ratings as \textit{positive feedback}(labeled with 1), and 1-3 star rating as \textit{negative feedback} (labeled with 0). Through feature embedding \cite{Wang2019DGL} and principal component analysis (PCA), dimension of each [user, item] pair (candidate context, \textit{i.e., behavior}) was reduced to 50, $i.e.,\ d=50$. For the candidate context set, we retained the original item for each interaction and then randomly sampled $9$ extra items from the entire item set (exclude the original item), so the number of candidates is $M=10$. In real recommendation, user-item interaction is sparse, meaning that not all of the items can be shown to the user and have feedback \cite{Zhang2021Counterfactual, Zhang2022Counteracting}. To overcome this issue, we trained two Matrix Factorization (MF) \cite{Koren2009Matrix} models using OL-data (denotes MF-OL) and BT-data  (denotes MF-BT) respectively for different baselines and IBCB model to serve as the simulated real environment. AUCs of the two MF models were both over $90\%$, which assures that the simulated environment can provide nearly realistic feedbacks of users. At step $b$ of $n$-th episode of both OL-data and BT-data phase, the simulated environment (MF-OL/MF-BT) received a selected candidate context $\bm s_{I_{n,b}}$ and a context set $\mathcal{S}_{n,b}$ consisted of all candidate contexts in this step, getting all candidate context's reward and sort out the candidate context $\bm s_{I_{\mathrm{max}}}$ with max reward $r_{I_{\mathrm{max}}}$. If selected candidate context $\bm s_{I_{n,b}}$ do not have the biggest reward compared to $\bm s_{I_{\mathrm{max}}}$, $i.e., r_{I_{n,b}} < r_{I_{\mathrm{max}}}$, reward feedback would be directly set as 0, otherwise would be set as $r_{I_{\mathrm{max}}}$. If $r_{I_{\mathrm{max}}}>0.5$, then the final reward feedback would be 1, otherwise would be 0. Adjustments for the ablation study period was also made similar with the synthetic dataset. Since duplicated data will not appeared in a large scale in real world, we just controlled the training rate similarly as that in synthetic dataset's setting.

Note that MF-OL is quite different compared to MF-BT (\textit{i.e.,} BT-data's distribution is different to OL-data's distribution), so experiments made on BT-data can directly provide the results of model's generalization ability in out-of-distribution data.

Finally, for OL-data in real-world dataset, we set numbers of episodes $N=20$, batch size $B=1000$, and for BT-data, we set $N=50$, $B=1000$ for precise measurement, considering real-world data's noise. Number of candidate contexts (actions) is $M=10$ at any single step of an episode in both OL-data and BT-data phase. 

\subsection{Evaluation Protocol}
\label{eval:pro}
\subsubsection{Online Train Log Fitness (OL-Fitness)}
\label{eval:pro:OL}OL-Fitness measures the alignment of various algorithms' learned parameters with the true reward environment's parameters. It involves using the learned parameters from IL algorithms (including IBCB) or backbone bandit algorithms to simulate a reward environment, where an expert interacts to gather data for online learning. The actions taken by the expert in this simulated environment are compared against those taken in the true reward environment, calculated as $\frac{1}{NB}\sum_{n=1}^{N}\sum_{b=1}^{B} \mathbb{I} (I_{n,b-\mathrm{estimated}} = I_{n,b-\mathrm{true}})$. Here, $I_{n,b-\mathrm{estimated}}$ denotes the action index from the estimated parameters in step $b$ of episode $n$ during the OL-data phase, and $I_{n,b-\mathrm{true}}$ is the action index from the true reward parameters under the same conditions. OL-data's candidate context sets, sequences and reward parameters keep same during the comparison by using fixed seeds. Note that BC algorithms cannot obtain reward parameters, and thus were not included in this comparison scope. But in \textit{dup-test} period, we used `train log behaviors fitness' to evaluate whether BC algorithm suffers from contradictory data.  A higher OL-Fitness indicates a closer match between the learned parameters and the true reward parameters.

\subsubsection{Batch Test Log Fitness (BT-Fitness)}
\label{eval:pro:BTF}BT-Fitness measures the alignment of parameters/models from various algorithms with those of the final expert policy. It is calculated by comparing the actions executed by IL algorithms against those of the expert policy during the BT-data phase, using the formula $\frac{1}{NB}\sum_{n=1}^{N}\sum_{b=1}^{B} \mathbb{I}(I_{n,b-\mathrm{IL}}=I_{n,b-\mathrm{expert}})$. Here, $I_{n,b-\mathrm{IL}}$ represents the action index executed by IL algorithms (including IBCB), while $I_{n,b-\mathrm{expert}}$ is the action index from the expert policy in step $b$ of episode $n$. BT-data's candidate context sets, sequences and reward parameters keep same during the comparison by using fixed seeds, and expert do not update its policy parameters in BT-data phase. Higher BT-Fitness indicates greater similarity to the expert policy's parameters. 

\subsubsection{Batch Test Average Reward (BT-AR)}
\label{eval:pro:BTAR}measures the reward performance of IL algorithms (including IBCB) against expert algorithms during the BT-data phase. It calculates average reward from executed actions after interacting with the BT-data environment, denoted as $\frac{1}{NB}\sum_{n=1}^{N}\sum_{b=1}^{B}R_{n,b}$. $R_{n,b}$ is binary reward for actions at step $b$ of episode $n$. Higher BT-AR indicates more effective actions in the BT-data environment, suggesting superior reward performance by the algorithm.

\subsubsection{Cumulative Fairness Regret (CFR)}
\label{eval:pro:CFR}CFR measures the similarity in fairness between the policies of various IL algorithms (including IBCB) and the bandit policy after the final episode (i.e., the experienced expert) with fairness optimization. Following definitions of fairness regret in \cite{wang2021fairness, gillen2018online} , we define CFR as $ \text{CFR} = \sum_{n=1}^{N}\sum_{b=1}^{B} {(D_{\mathrm{JS}}(\pi_\mathrm{IL}(\mathcal{S}_{n,b}) || \pi_\mathrm{expert}(\mathcal{S}_{n,b}))} + |R_{n,b-\mathrm{IL}} - R_{n,b-\mathrm{expert}}|), \ R_{n,b-\mathrm{IL}}, R_{n,b-\mathrm{expert}}\in \{0,1\}$, where $D_{\mathrm{JS}}(P||Q)$ represents the Jensen-Shannon divergence \cite{menendez1997jensen} between distributions $P$ and $Q$. $\pi_\mathrm{IL}(\mathcal{S}_{n,b})$ refers to the policy obtained by various IL algorithms (including IBCB), and $\pi_\mathrm{expert}(\mathcal{S}_{n,b})$ refers to the bandit policy after the final episode (i.e., the experienced expert). Both of the above are the normalized probabilities of each context which can be selected in the candidate context list $\mathcal{S}_{n,b}$ at step $b$ of episode $n$ in BT-data phase. $R_{n,b-\mathrm{IL}}$ and $R_{n,b-\mathrm{expert}}$ are executed action's observed reward at step $b$ of the $n$-th episode in BT-data phase from learned model and final expert respectively. It should be noted that CFR is only used in the BT-data phase, that is, when the parameters of all models have been fixed. A lower CFR indicates a closer similarity to the fairness of the final expert policy, that is, a more similar performance to the expert policy with fairness optimization. 

\subsection{Training Details}\label{exp:details:train}
We run all of the experiments on the server equipped with 16 Intel Xeon E5-2630 v4@2.20GHz cores, and during training, all of the baselines and IBCB only uses the CPU. Note that all baselines and IBCB were training on expert behavioral evolution history of Online Learning. Detailed settings of baselines and IBCB are shown as follows.
\subsubsection{Original Expert Policy Settings} \label{exp:details:train:expert}
\textbf{SBUCB}: We set the hyper parameter ($\alpha$ in step 5 of Algorithm~\ref{alg:CBB:bandit:TS}, \textit{i.e., exploration parameter}) of SBUCB to 0.4 for both synthetic and real-world ML-100K dataset experiments. Note that this parameter is only effective in Online Learning phase since expert (SBUCB) do not update its policy parameters (\textit{i.e., explore}) in Batch Test phase.

\textbf{SBTS}: We set the hyper parameter ($\alpha$ in step 7 of Algorithm~\ref{alg:CBB:bandit:TS}, \textit{i.e., exploration parameter}) of SBTS to 0.4 for synthetic dataset experiments and 0.05 for real-world ML-100K dataset experiments. Note that this parameter is only effective in Online Learning phase since expert (SBTS) do not update its policy parameters (\textit{i.e., explore}) in Batch Test phase. 

\subsubsection{Baseline Settings}\label{exp:details:train:base}
\textbf{BaseSVC}: We followed the setting recommended in $\mathrm{sklearn.svm.LinearSVC}$ and set the $\mathrm{max\_iteration}$ (\textit{i.e., the maximum number of iterations}) to 1,000,000 to ensure that the model can converge in the end.

\textbf{B-IRL}: Following \cite{Huyuk2022Inverse}, we have run Metropolis-Hastings algorithm for 10,000 iterations to obtain 1,000 samples on synthetic dataset (500 samples and 5,000 iterations for real-world ML-100K dataset) with intervals of 10 iterations between each sample after 10,000 burn-in iterations (5,000 for real-world ML-100K dataset). The final estimated parameters are formed by simply averaging all 1,000 samples. (500 for real-world ML-100K dataset).

\textbf{B-ICB}: We followed the setting used in \cite{Huyuk2022Inverse}'s work. We changed the setting of $\mathrm{compute\_rhox}$ period (\textit{i.e., computing the reward parameters}) in real-world ML-100K dataset experiments. Since that ML-100K's user embedding used one-hot encoding, which may make the final matrix $\bm \beta_n$ to concatenate in $\mathrm{compute\_rhox}$ be non-invertible, we replace the original $\mathrm{np.linalg.inv}$ method with $\mathrm{np.linalg.pinv}$ method to compute the pseudo inverse of $\bm \beta_n$, making sure experiment can be done without computational errors. And due to that $\bm \beta_n$ may be non-invertible, we also dismiss the $\mathrm{det}(\beta_n/\sigma^2)$ during iteration, where $\mathrm{det}(\bm A)$ means the determinant of matrix $\bm A$.

\textbf{Data Sampling Rate for B-ICB} denotes the proportion of training data B-ICB used for training since B-ICB cannot be run with large-scale training data. When the data sampling rate is $k \%$, we uniformly randomly sample $k \%$ of the original expert's training data as the training data (behavioral evolution history) for B-ICB. Data sampling rate was set to be 1\% to make sure that B-ICB can be trained in 100s on synthetic dataset experiments, and 0.5\% to make sure that B-ICB can be trained in 150s on real-world ML-100K dataset experiments. Also, in ablation study section, we used this setting for B-ICB.

\subsubsection{IBCB Settings}
We used OSQP solver to solve Eq.\eqref{eq:inverse:opt:constriant}'s quadratic problem(QP) as mentioned. Tolerance levels of OSQP was set to be $\epsilon_{abs}=8 \times 10^{-2}, \epsilon_{rel}=8 \times 10^{-2}$. If QP with the former setting cannot convergent, then setting would be modified as $\epsilon_{abs}=8 \times 10^{-1}, \epsilon_{rel}=8 \times 10^{-1}$. For real-world ML-100K experiments with SBTS backbone, setting was $\epsilon_{abs}=3 \times 10^{-1}, \epsilon_{rel}=3 \times 10^{-1}$.
Also, when expert's policy is SBTS, right-hand side of Eq.\eqref{eq:inverse:opt:constriant} should be set to a constant $\epsilon$ slightly greater than 0 to prevent the parameters learned by IBCB from being too close to the origin ($\bm 0_d$). Therefore, we set IBCB's $\epsilon$ to 0.01 in the experimental settings of the synthetic dataset and 0.001 in real-world ML-100K dataset when we assume expert's policy is SBTS.

\subsection{Experimental Results \& Analyses}
In this section, we compared several indicators between IBCB with other baselines. We used data sampling method to train B-ICB (namely B-ICB-S, where `-S' means `used sampled data') since B-ICB has not yet been adapted to batched version, so B-ICB can only be trained in environment that quantity of train data is not so large. 

All results were averaged over 5 different runs, and sampled the data over 5 different random seeds for each train data log (behavioral evolution history log) of expert on both synthetic and real-world ML-100K dataset. The source code has been available at \url{https://anonymous.4open.science/r/IBCB-F25F}.
\subsubsection{Experiments on Synthetic Dataset} 
\begin{table*}[!ht]
    \centering
    \setlength{\abovecaptionskip}{5pt}
    \setlength{\belowcaptionskip}{-2pt}
    \caption{{Reward parameters' estimation (\textbf{Upper Subtable}) and Train time (\textbf{Lower Subtable}) comparison on synthetic dataset, where $\mathrm{std}$ denotes the standard deviation of zero-mean Gaussian noise added to the final reward. Experts (SBUCB \& SBTS) do not need to train since they produce the train logs (behavioral evolution history logs) for baselines and IBCB.}}
    
    \scriptsize{
    \renewcommand{\arraystretch}{1.2}
    \begin{tabular}{l|c|c|c|c||l|c|c|c|c}
        \hline
        \multirow{2}{*}{Algorithm} & \multicolumn{4}{c||}{Online Train Log Fitness (OL-Fitness)} & \multirow{2}{*}{Algorithm} & \multicolumn{4}{c}{Online Train Log Fitness (OL-Fitness)}\\
        \cline{2-5} \cline{7-10} ~ & std=0 & std=0.03 & std=0.07 & std=0.10 &  ~ & std=0 & std=0.03 & std=0.07 & std=0.10  \\ 
        \hline \hline
        SBUCB & \textcolor[rgb]{0,0,0}{0.883±0.007} & \textcolor[rgb]{0,0,0}{0.879±0.016} & \textcolor[rgb]{0,0,0}{0.878±0.018} & \textcolor[rgb]{0,0,0}{0.874±0.019} &  SBTS & \textcolor[rgb]{0,0,0}{0.857±0.017} & \textcolor[rgb]{0,0,0}{0.876±0.012} & \textcolor[rgb]{0,0,0}{0.872±0.014} & \textcolor[rgb]{0,0,0}{0.871±0.008} \\ 
        \hline
        B-IRL & {0.827±0.012} & 0.808±0.039 & 0.806±0.037 & {0.802±0.036} &  B-IRL & {\textcolor[rgb]{0,0,0}{0.457±0.020}} & {\textcolor[rgb]{0,0,0}{0.468±0.029}} & {\textcolor[rgb]{0,0,0}{0.480±0.023}} & {\textcolor[rgb]{0,0,0}{0.476±0.039}}  \\
        B-ICB-S & 0.805±0.020 & 0.771±0.031 & 0.780±0.034 & 0.771±0.024 & B-ICB-S & 0.424±0.020 & 0.451±0.036 & 0.457±0.016 & 0.454±0.024  \\
        \hline
        \textbf{IBCB (Ours)} & \textbf{\textcolor[rgb]{0,0,0}{0.868±0.008}} & \textbf{\textcolor[rgb]{0,0,0}{0.858±0.036}} & \textbf{\textcolor[rgb]{0,0,0}{0.859±0.038}} & \textbf{\textcolor[rgb]{0,0,0}{0.861±0.029}} & \textbf{IBCB (Ours)} & \textbf{\textcolor[rgb]{0,0,0}{0.835±0.014}} & \textbf{\textcolor[rgb]{0,0,0}{0.856±0.009}} & \textbf{\textcolor[rgb]{0,0,0}{0.849±0.014}} & \textbf{\textcolor[rgb]{0,0,0}{0.849±0.009}}  \\ 
        \hline
    \end{tabular}}

    \vspace{1em}
    
    \scriptsize{
    \renewcommand{\arraystretch}{1.2}
    \begin{tabular}{l|c|c|c|c||l|c|c|c|c}
        \hline
        \multirow{2}{*}{Algorithm} & \multicolumn{4}{c||}{Train Time (sec.)} & \multirow{2}{*}{Algorithm} & \multicolumn{4}{c}{Train Time (sec.)}\\
        \cline{2-5} \cline{7-10} ~ & std=0 & std=0.03 & std=0.07 & std=0.10 & ~ & std=0 & std=0.03 & std=0.07 & std=0.10 \\ 
        \hline \hline
        SBUCB & \textcolor[rgb]{0,0,0}{/} & \textcolor[rgb]{0,0,0}{/} & \textcolor[rgb]{0,0,0}{/} & \textcolor[rgb]{0,0,0}{/} & SBTS & \textcolor[rgb]{0,0,0}{/} & \textcolor[rgb]{0,0,0}{/} & \textcolor[rgb]{0,0,0}{/} & \textcolor[rgb]{0,0,0}{/}  \\ 
        \hline
        B-IRL & {36.997±0.803} & {37.67±1.328} & {39.91±3.419} & {37.424±0.563} & B-IRL & {\textcolor[rgb]{0,0,0}{38.46±3.550}} & {\textcolor[rgb]{0,0,0}{36.419±0.187}} & {\textcolor[rgb]{0,0,0}{37.843±3.486}} & {\textcolor[rgb]{0,0,0}{43.469±6.228}}  \\ 
        BaseSVC & 116.86±20.63 & 116.39±26.58 & 132.18±19.86 & 117.25±19.47 & BaseSVC & 96.603±9.619 & 87.310±8.435 & 89.240±6.566 & 90.472±8.494 \\ 
        B-ICB-S & 86.365±0.827 & 87.599±4.489 & 86.85±1.162 & 85.713±0.676 & B-ICB-S & 91.166±1.164 & 89.193±1.477 & 89.892±1.162 & 89.483±1.585 \\ 
        \hline
        \textbf{IBCB (Ours)} & \textbf{\textcolor[rgb]{0,0,0}{0.815±0.045}} & \textbf{\textcolor[rgb]{0,0,0}{0.999±0.128}} & \textbf{\textcolor[rgb]{0,0,0}{0.866±0.137}} & \textbf{\textcolor[rgb]{0,0,0}{1.016±0.109}} & \textbf{IBCB (Ours)} & \textbf{\textcolor[rgb]{0,0,0}{0.778±0.035}} & \textbf{\textcolor[rgb]{0,0,0}{0.809±0.077}} & \textbf{\textcolor[rgb]{0,0,0}{0.772±0.034}} & \textbf{\textcolor[rgb]{0,0,0}{0.871±0.216}}  \\ 
        \hline
    \end{tabular}
    
     }
    
    \label{tab:real:tot:v2}
\end{table*}

\begin{table*}[!ht]
    \centering
    \setlength{\abovecaptionskip}{5pt}
    \setlength{\belowcaptionskip}{-2pt}
    \caption{{Expert policy (SBUCB) parameters' estimation (\textbf{Upper Subtable}), Expert policy (SBTS) parameters' estimation (\textbf{Lower Subtable}) on synthetic dataset. Expert policy produces same actions compared with itself, so Batch Test Log Fitness (BT-Fitness) of expert policy is 1.}}
    \scriptsize{
    \renewcommand{\arraystretch}{1.2}
    \begin{tabular}{l|c|c|c|c||l|c|c|c|c}
        \hline
        \multirow{2}{*}{Algorithm} & \multicolumn{4}{c||}{Batch Test Log Fitness (BT-Fitness)} & \multirow{2}{*}{Algorithm} & \multicolumn{4}{c}{Batch Test Average Reward (BT-AR)}\\
        \cline{2-5} \cline{7-10} ~ & std=0 & std=0.03 & std=0.07 & std=0.10 & ~ & std=0 & std=0.03 & std=0.07 & std=0.10 \\ 
        \hline \hline
        SBUCB & \textcolor[rgb]{0,0,0}{1} & \textcolor[rgb]{0,0,0}{1} & \textcolor[rgb]{0,0,0}{1} & \textcolor[rgb]{0,0,0}{1} & SBUCB & \textcolor[rgb]{0,0,0}{0.641±0.010} & \textcolor[rgb]{0,0,0}{0.629±0.028} & \textcolor[rgb]{0,0,0}{0.630±0.024} & \textcolor[rgb]{0,0,0}{0.622±0.024}  \\ 
        \hline
        B-IRL & {0.828±0.005} & {0.820±0.013} & {0.820±0.014} & {0.818±0.016} & B-IRL & {\textcolor[rgb]{0,0,0}{0.588±0.015}} & {\textcolor[rgb]{0,0,0}{0.571±0.037}} & {\textcolor[rgb]{0,0,0}{0.572±0.034}} & {\textcolor[rgb]{0,0,0}{0.561±0.034}}  \\ 
        BaseSVC & 0.791±0.008 & 0.781±0.020 & 0.781±0.020 & 0.777±0.020 & BaseSVC & 0.577±0.017 & 0.559±0.042 & 0.559±0.037 & 0.547±0.037 \\ 
        B-ICB-S & 0.800±0.032 & 0.767±0.055 & 0.780±0.045 & 0.776±0.022 & B-ICB-S & 0.581±0.006 & 0.553±0.024 & 0.560±0.025 & 0.552±0.025 \\ 
        \hline
        \textbf{IBCB (Ours)} & \textbf{\textcolor[rgb]{0,0,0}{0.972±0.011}} & \textbf{\textcolor[rgb]{0,0,0}{0.963±0.031}} & \textbf{\textcolor[rgb]{0,0,0}{0.964±0.028}} & \textbf{\textcolor[rgb]{0,0,0}{0.972±0.016}} & \textbf{IBCB (Ours)} & \textbf{\textcolor[rgb]{0,0,0}{0.647±0.012}} & \textbf{\textcolor[rgb]{0,0,0}{0.638±0.026}} & \textbf{\textcolor[rgb]{0,0,0}{0.639±0.019}} & \textbf{\textcolor[rgb]{0,0,0}{0.628±0.022}}  \\ 
        \hline
    \end{tabular}
    }
    \vspace{1em}

    \scriptsize{
    \renewcommand{\arraystretch}{1.2}
    \begin{tabular}{l|c|c|c|c||l|c|c|c|c}
        \hline
        \multirow{2}{*}{Algorithm} & \multicolumn{4}{c||}{Batch Test Log Fitness (BT-Fitness)} & \multirow{2}{*}{Algorithm} & \multicolumn{4}{c}{Batch Test Average Reward (BT-AR)}\\
        \cline{2-5} \cline{7-10} ~ & std=0 & std=0.03 & std=0.07 & std=0.10 & ~ & std=0 & std=0.03 & std=0.07 & std=0.10 \\ 
        \hline \hline
        SBTS & \textcolor[rgb]{0,0,0}{1} & \textcolor[rgb]{0,0,0}{1} & \textcolor[rgb]{0,0,0}{1} & \textcolor[rgb]{0,0,0}{1} & SBTS & \textcolor[rgb]{0,0,0}{0.650±0.020} & \textcolor[rgb]{0,0,0}{0.662±0.025} & \textcolor[rgb]{0,0,0}{0.661±0.025} & \textcolor[rgb]{0,0,0}{0.656±0.024}  \\ 
        \hline
        B-IRL & {0.816±0.007} & {0.820±0.015} & {0.816±0.008} & {0.816±0.006} & B-IRL & {0.595±0.026} & {0.615±0.028} & {0.611±0.031} & {0.605±0.030}  \\ 
        BaseSVC & 0.789±0.011 & 0.807±0.009 & 0.801±0.015 & 0.798±0.012 & BaseSVC & 0.586±0.032 & 0.611±0.035 & 0.607±0.037 & 0.600±0.034 \\ 
        B-ICB-S & 0.740±0.059 & 0.742±0.089 & 0.741±0.074 & 0.752±0.076 & B-ICB-S & 0.571±0.021 & 0.589±0.022 & 0.588±0.021 & 0.584±0.023 \\ 
        \hline
        \textbf{IBCB (Ours)} & \textbf{\textcolor[rgb]{0,0,0}{0.963±0.012}} & \textbf{\textcolor[rgb]{0,0,0}{0.958±0.010}} & \textbf{\textcolor[rgb]{0,0,0}{0.958±0.006}} & \textbf{\textcolor[rgb]{0,0,0}{0.958±0.005}} & \textbf{IBCB (Ours)} & \textbf{\textcolor[rgb]{0,0,0}{0.651±0.027}} & \textbf{\textcolor[rgb]{0,0,0}{0.667±0.027}} & \textbf{\textcolor[rgb]{0,0,0}{0.666±0.026}} & \textbf{\textcolor[rgb]{0,0,0}{0.658±0.029}}  \\ 
        \hline
    \end{tabular}
    }
    
    \label{tab:bt:bandit:tot:v2}
\end{table*}
Table~\ref{tab:real:tot:v2} display OL-Fitness and training time for experts (SBUCB \& SBTS), baselines, and IBCB. IBCB outperforms others with higher OL-Fitness and faster training, thanks to its optimal reward parameter estimation and OSQP solver's efficiency in managing large-scale QP problems with nearly linear complexity. BaseSVC trains slower due to its complexity near to quadratic, while B-IRL requires many Metropolis-Hastings iterations for better performance. B-ICB's quadratic complexity slows its training, despite sampling data.\\
Table~\ref{tab:bt:bandit:tot:v2} show that IBCB also achieves higher BT-Fitness and BT-AR, highlighting the effectiveness of its reward imputation method for learning from the expert's behavioral evolution history. This confirms IBCB's ability to learn the expert policy's parameters effectively and efficiently, even in out-of-distribution (OOD) scenarios as seen in Table~\ref{tab:bt:OOD:bandit:tot} in Section~\ref{app:exp:details:OOD}, demonstrating its robustness to distribution shifts.

\subsubsection{Experiments on Real-World Dataset  (ML-100K Dataset)} 
\begin{table*}[!ht]
    \centering
    \setlength{\abovecaptionskip}{5pt}
    \setlength{\belowcaptionskip}{-5pt}
    \caption{{Reward parameters' estimation , Train time comparison and Expert policy parameters' estimation on ML-100K dataset. Experts do not need to train since they produce train logs (behavioral evolution history logs) for baselines and IBCB. BaseSVC cannot obtain reward parameters, so it was excluded from OL-Fitness comparison.  Expert policy produces same actions compared with itself, so BT-Fitness of expert policy is 1.}}
    
    \scriptsize{
    \renewcommand{\arraystretch}{1.1}
    \begin{tabular}{l|c|c|c|c||l|c|c|c|c}
        \hline
        Algorithm & OL-Fitness & Train Time (sec.) & BT-Fitness & BT-AR & Algorithm & OL-Fitness & Train Time (sec.) & BT-Fitness & BT-AR \\
        \hline \hline
        SBUCB & \textcolor[rgb]{0,0,0}{0.891±0.006} & \textcolor[rgb]{0,0,0}{/} & \textcolor[rgb]{0,0,0}{1} & \textcolor[rgb]{0,0,0}{0.193±0.005} & SBTS & \textcolor[rgb]{0,0,0}{0.868±0.015} & \textcolor[rgb]{0,0,0}{/} & \textcolor[rgb]{0,0,0}{1} & \textcolor[rgb]{0,0,0}{0.200±0.005}  \\ 
        \hline
        B-IRL & {0.830±0.010} & {43.669±1.282} & {0.864±0.008} & {0.189±0.005} &  B-IRL & {0.851±0.015} & {47.005±4.311} & \textcolor[rgb]{0,0,0}{0.870±0.030} & {0.199±0.007} \\ 
        BaseSVC & / & 162.767±12.135 & 0.804±0.012 & {0.189±0.005} & BaseSVC & / & 192.83±33.199 & 0.801±0.019 & 0.198±0.007  \\ 
        B-ICB-S & 0.746±0.032 & 147.474±4.633 & 0.767±0.028 & 0.187±0.007 & B-ICB-S & 0.721±0.030 & 145.441±5.916 & 0.759±0.035 & 0.196±0.004 \\
        \hline
        \textbf{IBCB (Ours)} & \textbf{\textcolor[rgb]{0,0,0}{0.862±0.023}} & \textbf{\textcolor[rgb]{0,0,0}{4.036±0.111}} & \textbf{\textcolor[rgb]{0,0,0}{0.925±0.019}} & \textbf{\textcolor[rgb]{0,0,0}{0.192±0.005}} & \textbf{IBCB (Ours)} & \textbf{\textcolor[rgb]{0,0,0}{0.869±0.021}} & \textbf{\textcolor[rgb]{0,0,0}{8.830±1.825}} & {\textbf{\textcolor[rgb]{0,0,0}{0.956±0.006}}} & \textbf{\textcolor[rgb]{0,0,0}{0.201±0.005}} \\ 
        \hline
    \end{tabular}
    
    }

    
    \label{tab:ML:real:tot:v2}
\end{table*}
Results on ML-100K dataset of different indicators are reported in Table~\ref{tab:ML:real:tot:v2}. Similar to the results on synthetic dataset, IBCB achieved significantly higher performance than other baselines in all indicators. 
From the outstanding results of training speed and average reward, we concluded that IBCB can also adapt to learning real-world behavioral evolution history generated by expert with great efficiency.

\subsubsection{Experiments on Fairness-Aware Expert Limitation} 
In the original fairness-aware experiment, we set the \textbf{top-$k$} with $k = 2$, which means we choose the final action from the \textbf{top-$2$} based on normalized probabilities. 
\begin{table*}[!htbp]
    \centering
    \setlength{\abovecaptionskip}{5pt}
    \setlength{\belowcaptionskip}{-2pt}
    \caption{{Fairness comparison (\textbf{Upper Subtable}), Fairness-aware train time comparison (\textbf{Lower Subtable}) on synthetic dataset (CFR definition can be found in Section \ref{eval:pro:CFR}). {Expert selects action from \textbf{top-$\bm 2$} actions with proportion of its learned probability distribution through all actions. Expert is defined as the fairest policy from the CFR metric, so its CFR is always 0.} Fairness-aware experts (SBUCB \& SBTS) do not need to train since they produce the train logs (behavioral evolution history logs) for baselines and IBCB.}}
    \scriptsize{
    \renewcommand{\arraystretch}{1.2}
    \begin{tabular}{l|c|c|c|c||l|c|c|c|c}
        \hline
        \multirow{2}{*}{Algorithm} & \multicolumn{4}{c||}{Cumulative Fairness Regret (CFR)} & \multirow{2}{*}{Algorithm} & \multicolumn{4}{c}{Cumulative Fairness Regret (CFR)}\\
        \cline{2-5} \cline{7-10} ~ & std=0 & std=0.03 & std=0.07 & std=0.10 & ~ & std=0 & std=0.03 & std=0.07 & std=0.10 \\ 
        \hline \hline
        SBUCB                & 0                       & 0                         & 0                         & 0                         & SBTS                 & 0                       & 0                        & 0                         & 0                        \\
        \hline
        B-IRL                & 2175.0±159.1         & 2330.0±235.8           & 2347.2±305.4            & 2449.0±344.1           & B-IRL                & 2106.5±136.6         & 2080.7±236.3          & 2102.9±227.4           & 2136.9±192.1           \\
        BaseSVC              & 2027.5±184.2        & 2218.2±342.6           & 2248.0±402.2           & 2367.8±449.9           & BaseSVC              & 1848.8±232.4          & 1776.9±427.0           & 1763.8±391.5           & 1841.2±344.7           \\
        B-ICB-S              & 2162.3±205.5         & 2661.7±169.6           & 2507.1±240.8           & 2437.6±175.9           & B-ICB-S              & 2679.5±405.6        & 2922.5±451.3         & 2660.4±305.1           & 2800.2±197.7          \\
        \hline
        \textbf{IBCB (ours)} & \textbf{1609.6±92.6} & \textbf{1475.8±127.2} & \textbf{1353.9±132.3} & \textbf{1409.8±205.8} & \textbf{IBCB (ours)} & \textbf{1263.2±104.3} & \textbf{1262.1±198.3} & \textbf{1287.0±183.3} & \textbf{1257.8±117.3} \\
        \hline
    \end{tabular}
    
    }
    \vspace{1em}

    \scriptsize{
    \renewcommand{\arraystretch}{1.2}
    \begin{tabular}{l|c|c|c|c||l|c|c|c|c}
        \hline
        \multirow{2}{*}{Algorithm} & \multicolumn{4}{c||}{Train Time (sec.)} & \multirow{2}{*}{Algorithm} & \multicolumn{4}{c}{Train Time (sec.)}\\
        \cline{2-5} \cline{7-10} ~ & std=0 & std=0.03 & std=0.07 & std=0.10 & ~ & std=0 & std=0.03 & std=0.07 & std=0.10 \\ 
        \hline \hline
        SBUCB & \textcolor[rgb]{0,0,0}{/} & \textcolor[rgb]{0,0,0}{/} & \textcolor[rgb]{0,0,0}{/} & \textcolor[rgb]{0,0,0}{/} & SBTS & \textcolor[rgb]{0,0,0}{/} & \textcolor[rgb]{0,0,0}{/} & \textcolor[rgb]{0,0,0}{/} & \textcolor[rgb]{0,0,0}{/}  \\ 
        \hline
        B-IRL & {41.466±2.898} & {42.802±10.953} & {40.348±2.888} & {42.361±4.737} & B-IRL & {\textcolor[rgb]{0,0,0}{38.332±4.622}} & {\textcolor[rgb]{0,0,0}{35.703±0.101}} & {\textcolor[rgb]{0,0,0}{35.867±0.287}} & {\textcolor[rgb]{0,0,0}{39.226±4.344}}  \\ 
        BaseSVC & 210.83±37.72 & 187.31±19.48 & 181.46±21.53 & 193.40±14.25 & BaseSVC & 152.30±13.81 & 147.73±15.27 & 146.71±8.414 & 145.46±7.936 \\ 
        B-ICB-S & 90.922±4.373 & 89.637±4.892 & 88.908±2.650 & 90.525±2.571 & B-ICB-S & 85.615±1.072 & 85.511±2.658 & 84.919±1.306 & 86.802±1.530 \\ 
        \hline
        \textbf{IBCB (Ours)} & \textbf{\textcolor[rgb]{0,0,0}{1.370±0.238}} & \textbf{\textcolor[rgb]{0,0,0}{1.308±0.198}} & \textbf{\textcolor[rgb]{0,0,0}{1.403±0.243}} & \textbf{\textcolor[rgb]{0,0,0}{1.191±0.175}} & \textbf{IBCB (Ours)} & \textbf{\textcolor[rgb]{0,0,0}{0.889±0.015}} & \textbf{\textcolor[rgb]{0,0,0}{0.916±0.028}} & \textbf{\textcolor[rgb]{0,0,0}{0.898±0.020}} & \textbf{\textcolor[rgb]{0,0,0}{0.937±0.079}}  \\ 
        \hline
    \end{tabular}
    
    }
    \label{tab:fair:top:2:tot}
\end{table*}

\begin{table*}[!ht]
    \centering
    \setlength{\abovecaptionskip}{5pt}
    \setlength{\belowcaptionskip}{-2pt}
    \caption{{Fairness-aware} expert policy (SBUCB) parameters' estimation (\textbf{Upper Subtable}) and {Fairness-aware} expert policy (SBTS) parameters' estimation (\textbf{Lower Subtable}) on synthetic dataset. {Expert selects action from \textbf{top-$\bm 2$} actions with proportion of its learned probability distribution through all actions.} Expert policy produces same actions compared with itself, so Batch Test Log Fitness of expert policy is 1.}
    \scriptsize{
    \renewcommand{\arraystretch}{1.2}
    \begin{tabular}{l|c|c|c|c||l|c|c|c|c}
        \hline
        \multirow{2}{*}{Algorithm} & \multicolumn{4}{c||}{Batch Test Log Fitness} & \multirow{2}{*}{Algorithm} & \multicolumn{4}{c}{Batch Test Average Reward}\\
        \cline{2-5} \cline{7-10} ~ & std=0 & std=0.03 & std=0.07 & std=0.10 & ~ & std=0 & std=0.03 & std=0.07 & std=0.10 \\ 
        \hline \hline
        SBUCB & \textcolor[rgb]{0,0,0}{1} & \textcolor[rgb]{0,0,0}{1} & \textcolor[rgb]{0,0,0}{1} & \textcolor[rgb]{0,0,0}{1} & SBUCB & \textcolor[rgb]{0,0,0}{0.589±0.008} & \textcolor[rgb]{0,0,0}{0.583±0.030} & \textcolor[rgb]{0,0,0}{0.579±0.029} & \textcolor[rgb]{0,0,0}{0.568±0.032}  \\ 
        \hline
        B-IRL & {0.445±0.003} & {0.431±0.007} & {0.432±0.010} & {0.430±0.014} & B-IRL & {\textcolor[rgb]{0,0,0}{0.536±0.012}} & {\textcolor[rgb]{0,0,0}{0.524±0.039}} & {\textcolor[rgb]{0,0,0}{0.519±0.040}} & {\textcolor[rgb]{0,0,0}{0.504±0.045}}  \\ 
        BaseSVC & 0.446±0.004 & 0.439±0.009 & 0.439±0.014 & 0.434±0.015 & BaseSVC & 0.530±0.013 & 0.516±0.041 & 0.511±0.042 & 0.495±0.047 \\ 
        B-ICB-S & 0.431±0.014 & 0.417±0.008 & 0.419±0.012 & 0.421±0.007 & B-ICB-S & 0.543±0.005 & 0.512±0.025 & 0.515±0.024 & 0.508±0.027 \\ 
        \hline
        \textbf{IBCB (Ours)} & \textbf{\textcolor[rgb]{0,0,0}{0.455±0.003}} & \textbf{\textcolor[rgb]{0,0,0}{0.461±0.008}} & \textbf{\textcolor[rgb]{0,0,0}{0.464±0.008}} & \textbf{\textcolor[rgb]{0,0,0}{0.464±0.012}} & \textbf{IBCB (Ours)} & \textbf{\textcolor[rgb]{0,0,0}{0.609±0.011}} & \textbf{\textcolor[rgb]{0,0,0}{0.600±0.025}} & \textbf{\textcolor[rgb]{0,0,0}{0.591±0.030}} & \textbf{\textcolor[rgb]{0,0,0}{0.585±0.033}}  \\ 
        \hline
    \end{tabular}
    }

    \vspace{1em}

    \scriptsize{
    \renewcommand{\arraystretch}{1.2}
    \begin{tabular}{l|c|c|c|c||l|c|c|c|c}
        \hline
        \multirow{2}{*}{Algorithm} & \multicolumn{4}{c||}{Batch Test Log Fitness} & \multirow{2}{*}{Algorithm} & \multicolumn{4}{c}{Batch Test Average Reward}\\
        \cline{2-5} \cline{7-10} ~ & std=0 & std=0.03 & std=0.07 & std=0.10 & ~ & std=0 & std=0.03 & std=0.07 & std=0.10 \\ 
        \hline \hline
        SBTS & \textcolor[rgb]{0,0,0}{1} & \textcolor[rgb]{0,0,0}{1} & \textcolor[rgb]{0,0,0}{1} & \textcolor[rgb]{0,0,0}{1} & SBTS & \textcolor[rgb]{0,0,0}{0.608±0.023} & \textcolor[rgb]{0,0,0}{0.614±0.033} & \textcolor[rgb]{0,0,0}{0.612±0.029} & \textcolor[rgb]{0,0,0}{0.606±0.027}  \\ 
        \hline
        B-IRL & {0.448±0.005} & {0.444±0.009} & {0.444±0.006} & {0.442±0.007} & B-IRL & {0.558±0.029} & {0.565±0.044} & {0.563±0.039} & {0.555±0.035}  \\ 
        BaseSVC & 0.451±0.007 & 0.452±0.007 & 0.449±0.005 & 0.449±0.008 & BaseSVC & 0.553±0.030 & 0.562±0.047 & 0.560±0.042 & 0.552±0.037 \\ 
        B-ICB-S & 0.409±0.023 & 0.396±0.027 & 0.406±0.017 & 0.402±0.013 & B-ICB-S & 0.536±0.022 & 0.530±0.018 & 0.541±0.016 & 0.529±0.022 \\ 
        \hline
        \textbf{IBCB (Ours)} & \textbf{\textcolor[rgb]{0,0,0}{0.472±0.006}} & \textbf{\textcolor[rgb]{0,0,0}{0.473±0.008}} & \textbf{\textcolor[rgb]{0,0,0}{0.473±0.007}} & \textbf{\textcolor[rgb]{0,0,0}{0.476±0.003}} & \textbf{IBCB (Ours)} & \textbf{\textcolor[rgb]{0,0,0}{0.617±0.024}} & \textbf{\textcolor[rgb]{0,0,0}{0.622±0.035}} & \textbf{\textcolor[rgb]{0,0,0}{0.621±0.033}} & \textbf{\textcolor[rgb]{0,0,0}{0.616±0.028}}  \\ 
        \hline
    \end{tabular}
    }
    \label{tab:fair:top:2:bt:bandit:tot}
    
\end{table*}

The experimental results in Table~\ref{tab:fair:top:2:tot} and Table~\ref{tab:fair:top:2:bt:bandit:tot} 
demonstrate that IBCB effectively learns expert parameters in complex, fairness-oriented recommendation settings. With a significantly lower CFR compared to the baseline and well-managed training time, IBCB achieves superior BT-Fitness and BT-AR, showing its capacity to uphold high recommendation quality while ensuring fairness. 
IBCB's exceptional performance indicates its robust generalization in learning fair expert parameters and its framework's adaptability in addressing BCB-related challenges.

\subsection{Ablation Study}
In this section, we explore extreme scenarios faced by imitation learning (IL) algorithms and IBCB, including delayed updates to training logs, which provides only partial access to the expert's behavioral history (e.g., the first half of the data), and the emergence of out-of-distribution or contradictory data. Since hyperparameter $\alpha$ of the original expert policy is inaccessible to IBCB, we also examine the impact of different $\alpha$ settings in IBCB. Ablation Study's results on synthetic dataset are detailed in Section~\ref{app:exp:details:OOD}--Section~\ref{app:exp:details:ab:param}. Also, comprehensive results and analysis of IBCB's hyperparameter tuning and ablation study on ML-100K dataset are detailed in Section~\ref{app:exp:details:ab:ML} and Section~\ref{app:exp:details:ab:ML:extra}.
\subsubsection{Out-of-distribution Data Experiments on Synthetic Dataset}
\label{app:exp:details:OOD}
\begin{table*}[!ht]
    \centering
    \setlength{\abovecaptionskip}{5pt}
    \setlength{\belowcaptionskip}{5pt}
    \caption{{Expert policy (SBUCB) parameters' estimation (\textbf{Upper Subtable}) and Expert policy (SBTS) parameters' estimation (\textbf{Lower Subtable}) for \textbf{Out-of-distribution data} on synthetic dataset. Expert policy produces same actions compared with itself, so BT-Fitness of expert policy is 1.}}
    \renewcommand{\arraystretch}{1.2}
    \scriptsize{\begin{tabular}{l|c|c|c|c||l|c|c|c|c}
        \hline
        \multirow{2}{*}{Algorithm} & \multicolumn{4}{c||}{Batch Test Log Fitness (OOD)} & \multirow{2}{*}{Algorithm} & \multicolumn{4}{c}{Batch Test Average Reward (OOD)}\\
        \cline{2-5} \cline{7-10} ~ & std=0 & std=0.03 & std=0.07 & std=0.10 & ~ & std=0 & std=0.03 & std=0.07 & std=0.10 \\ 
        \hline \hline
        SBUCB & \textcolor[rgb]{0,0,0}{1} & \textcolor[rgb]{0,0,0}{1} & \textcolor[rgb]{0,0,0}{1} & \textcolor[rgb]{0,0,0}{1} & SBUCB & \textcolor[rgb]{0,0,0}{0.709±0.015} & \textcolor[rgb]{0,0,0}{0.695±0.036} & \textcolor[rgb]{0,0,0}{0.694±0.031} & \textcolor[rgb]{0,0,0}{0.684±0.030} \\ 
        \hline
        B-IRL & {0.802±0.005} & {0.793±0.013} & {0.795±0.015} & {0.793±0.016} & B-IRL & {\textcolor[rgb]{0,0,0}{0.642±0.019}} & {\textcolor[rgb]{0,0,0}{0.622±0.047}} & {\textcolor[rgb]{0,0,0}{0.622±0.042}} & {\textcolor[rgb]{0,0,0}{0.608±0.041}} \\ 
        BaseSVC & 0.764±0.009 & 0.756±0.021 & 0.756±0.022 & 0.750±0.023 & BaseSVC & 0.627±0.021 & 0.607±0.052 & 0.606±0.046 & 0.591±0.046  \\ 
        B-ICB-S & 0.775±0.039 & 0.738±0.065 & 0.755±0.050 & 0.754±0.025 & B-ICB-S & 0.632±0.008 & 0.600±0.030 & 0.607±0.032 & 0.596±0.032  \\ 
        \hline
        \textbf{IBCB (Ours)} & \textbf{\textcolor[rgb]{0,0,0}{0.970±0.012}} & \textbf{\textcolor[rgb]{0,0,0}{0.960±0.036}} & \textbf{\textcolor[rgb]{0,0,0}{0.959±0.034}} & \textbf{\textcolor[rgb]{0,0,0}{0.969±0.02}} & \textbf{IBCB (Ours)} & \textbf{\textcolor[rgb]{0,0,0}{0.717±0.017}} & \textbf{\textcolor[rgb]{0,0,0}{0.707±0.034}} & \textbf{\textcolor[rgb]{0,0,0}{0.706±0.024}} & \textbf{\textcolor[rgb]{0,0,0}{0.693±0.028}}   \\ 
        \hline
    \end{tabular}
    }

    \vspace{1em}

    \renewcommand{\arraystretch}{1.2}
    \scriptsize{\begin{tabular}{l|c|c|c|c||l|c|c|c|c}
        \hline
        \multirow{2}{*}{Algorithm} & \multicolumn{4}{c||}{Batch Test Log Fitness (OOD)} & \multirow{2}{*}{Algorithm} & \multicolumn{4}{c}{Batch Test Average Reward (OOD)}\\
        \cline{2-5} \cline{7-10} ~ & std=0 & std=0.03 & std=0.07 & std=0.10 & ~ & std=0 & std=0.03 & std=0.07 & std=0.10 \\ 
        \hline \hline
        SBTS & \textcolor[rgb]{0,0,0}{1} & \textcolor[rgb]{0,0,0}{1} & \textcolor[rgb]{0,0,0}{1} & \textcolor[rgb]{0,0,0}{1} &  SBTS & \textcolor[rgb]{0,0,0}{0.719±0.028} & \textcolor[rgb]{0,0,0}{0.739±0.032} & \textcolor[rgb]{0,0,0}{0.735±0.032} & \textcolor[rgb]{0,0,0}{0.729±0.032}   \\ 
        \hline
        B-IRL & {0.792±0.005} & {0.795±0.010} & {0.793±0.006} & {0.793±0.005} & B-IRL & {0.650±0.034} & {0.676±0.037} & {0.671±0.039} & {0.663±0.037}   \\ 
        BaseSVC & 0.767±0.015 & 0.782±0.017 & 0.780±0.019 & 0.776±0.014 & BaseSVC & 0.640±0.041 & 0.670±0.045 & 0.666±0.047 & 0.658±0.044  \\ 
        B-ICB-S & 0.714±0.065 & 0.710±0.096 & 0.712±0.081 & 0.724±0.086 & B-ICB-S & 0.620±0.028 & 0.643±0.028 & 0.642±0.026 & 0.637±0.027  \\ 
        \hline
        \textbf{IBCB (Ours)} & \textbf{\textcolor[rgb]{0,0,0}{0.965±0.013}} & \textbf{\textcolor[rgb]{0,0,0}{0.962±0.009}} & \textbf{\textcolor[rgb]{0,0,0}{0.961±0.003}} & \textbf{\textcolor[rgb]{0,0,0}{0.962±0.005}} & \textbf{IBCB (Ours)} & \textbf{\textcolor[rgb]{0,0,0}{0.722±0.035}} & \textbf{\textcolor[rgb]{0,0,0}{0.744±0.034}} & \textbf{\textcolor[rgb]{0,0,0}{0.743±0.033}} & \textbf{\textcolor[rgb]{0,0,0}{0.732±0.038}}   \\ 
        \hline
    \end{tabular}
    }
    \label{tab:bt:OOD:bandit:tot}
\end{table*}

As shown in Table~\ref{tab:bt:OOD:bandit:tot}, IBCB still outperforms other IL baselines in expert policy's parameter estimation on out-of-distribution (OOD) data.
\subsubsection{Limited Training logs of Experts on Synthetic Dataset}
\label{app:exp:details:limited:training}
\begin{table}[!ht]
    \centering
    \setlength{\abovecaptionskip}{3pt}
    \setlength{\belowcaptionskip}{-3pt}
    \caption{Reward parameters' estimation on synthetic dataset (using \textbf{first half}of expert's behavior evolution history logs). BaseSVC cannot obtain reward parameters, so it was excluded from OL-Fitness comparison.}
    \renewcommand{\arraystretch}{1.2}
    \scriptsize{\begin{tabular}{l|c|c|c|c}
        \hline
        \multirow{2}{*}{Algorithm} & \multicolumn{4}{c}{Online Train Log Fitness (Half Train)} \\
        \cline{2-5} ~ & std=0 & std=0.03 & std=0.07 & std=0.10  \\ 
        \hline \hline
        SBUCB & \textcolor[rgb]{0,0,0}{0.883±0.007} & \textcolor[rgb]{0,0,0}{0.879±0.016} & \textcolor[rgb]{0,0,0}{0.878±0.018} & \textcolor[rgb]{0,0,0}{0.874±0.019}  \\
        \hline
        B-IRL & {0.752±0.018} & {0.729±0.051} & {0.728±0.050} & {0.721±0.047}  \\ 
        B-ICB-S & 0.730±0.039 & 0.727±0.045 & 0.721±0.063 & 0.717±0.041  \\
        \hline
        \textbf{IBCB (Ours)} & \textbf{\textcolor[rgb]{0,0,0}{0.895±0.011}} & \textbf{\textcolor[rgb]{0,0,0}{0.894±0.016}} & \textbf{\textcolor[rgb]{0,0,0}{0.892±0.024}} & \textbf{\textcolor[rgb]{0,0,0}{0.886±0.017}}  \\ 
        \hline
    \end{tabular}
    }  
    \label{tab:ab:ce:real:env:val:tot}
\end{table}
\begin{table}[!ht]
    \centering
    \setlength{\abovecaptionskip}{3pt}
    \setlength{\belowcaptionskip}{-3pt}
    \caption{{Expert policy parameters' estimation on synthetic dataset (using \textbf{first half} of expert's behavioral evolution history logs). Expert policy produces same actions compared with itself, so BT-Fitness of expert policy is 1.}}
    \renewcommand{\arraystretch}{1.2}
    \scriptsize{\begin{tabular}{l|c|c|c|c}
        \hline
        \multirow{2}{*}{Algorithm} & \multicolumn{4}{c}{Batch Test Log Fitness (Half Train)} \\
        \cline{2-5} ~ & std=0 & std=0.03 & std=0.07 & std=0.10 \\ 
        \hline \hline
        SBUCB & \textcolor[rgb]{0,0,0}{1} & \textcolor[rgb]{0,0,0}{1} & \textcolor[rgb]{0,0,0}{1} & \textcolor[rgb]{0,0,0}{1} \\ 
        \hline
        B-IRL & {0.718±0.012} & {0.699±0.031} & {0.702±0.035} & {0.694±0.037} \\ 
        BaseSVC & 0.656±0.019 & 0.638±0.039 & 0.638±0.043 & 0.627±0.045 \\ 
        B-ICB-S & 0.695±0.054 & 0.707±0.069 & 0.695±0.084 & 0.688±0.043 \\ 
        \hline
        \textbf{IBCB (Ours)} & \textbf{\textcolor[rgb]{0,0,0}{0.891±0.019}} & \textbf{\textcolor[rgb]{0,0,0}{0.919±0.030}} & \textbf{\textcolor[rgb]{0,0,0}{0.890±0.031}} & \textbf{\textcolor[rgb]{0,0,0}{0.890±0.023}} \\ 
        \hline
    \end{tabular}
    }

    \vspace{1em}

    \renewcommand{\arraystretch}{1.2}
    \scriptsize{\begin{tabular}{l|c|c|c|c}
        \hline
        \multirow{2}{*}{Algorithm} & \multicolumn{4}{c}{Batch Test Average Reward (Half Train)}\\
        \cline{2-5} ~ & std=0 & std=0.03 & std=0.07 & std=0.10 \\ 
        \hline \hline
        SBUCB & \textcolor[rgb]{0,0,0}{0.641±0.010} & \textcolor[rgb]{0,0,0}{0.629±0.028} & \textcolor[rgb]{0,0,0}{0.630±0.024} & \textcolor[rgb]{0,0,0}{0.622±0.024}  \\ 
        \hline
        B-IRL & 0.545±0.020 & 0.522±0.047 & 0.522±0.046 & 0.510±0.046  \\ 
        BaseSVC & 0.520±0.023 & 0.497±0.053 & 0.497±0.052 & 0.482±0.051 \\ 
        B-ICB-S & {0.549±0.019} & {0.538±0.024} & {0.540±0.030} & {0.529±0.033} \\ 
        \hline
        \textbf{IBCB (Ours)} & \textbf{\textcolor[rgb]{0,0,0}{0.610±0.015}} & \textbf{\textcolor[rgb]{0,0,0}{0.608±0.025}} & \textbf{\textcolor[rgb]{0,0,0}{0.597±0.036}} & \textbf{\textcolor[rgb]{0,0,0}{0.588±0.030}}  \\ 
        \hline
    \end{tabular}
    }
    \label{tab:ab:ce:bt:bandit:SBUCB:tot}
\end{table}
From the results in Table~\ref{tab:ab:ce:real:env:val:tot} and Table~\ref{tab:ab:ce:bt:bandit:SBUCB:tot} we find that, when all IL baselines and IBCB only have the access to the first half of expert's behavioral evolution history logs, \textit{i.e., novice expert's evolution history}, IBCB can have better performance, compared with the degree of decline in all algorithms from Table~\ref{tab:real:tot:v2} and Table~\ref{tab:bt:bandit:tot:v2}.
\subsubsection{Extra Fairness-aware Expert Experiments on Synthetic Dataset}
\label{app:exp:details:extra:fair}
\begin{table*}[!htbp]
    \centering
    \setlength{\abovecaptionskip}{5pt}
    \setlength{\belowcaptionskip}{-2pt}
    \caption{{Cumulative Fairness Regret (CFR) comparison (\textbf{Upper Subtable}) and Train time comparison (\textbf{Lower Subtable}) on synthetic dataset. {Expert selects action from \textbf{top-$\bm 3$} actions with proportion of its learned probability distribution through all actions. Expert is defined as the fairest policy from the CFR metric, so its CFR is always 0. Fairness-aware experts (SBUCB \& SBTS) do not need to train since they produce the train logs (behavioral evolution history logs) for baselines and IBCB.}}}
    \renewcommand{\arraystretch}{1.2}
    \scriptsize{\begin{tabular}{l|c|c|c|c||l|c|c|c|c}
        \hline
        \multirow{2}{*}{Algorithm} & \multicolumn{4}{c||}{Cumulative Fairness Regret} & \multirow{2}{*}{Algorithm} & \multicolumn{4}{c}{Cumulative Fairness Regret}\\
        \cline{2-5} \cline{7-10} ~ & std=0 & std=0.03 & std=0.07 & std=0.10 & ~ & std=0 & std=0.03 & std=0.07 & std=0.10 \\ 
        \hline \hline
        SBUCB                & 0                         & 0                         & 0                       & 0 & SBTS & 0  & 0 & 0  & 0  \\
        \hline
        B-IRL                & 2713.9±230.2          & 2882.3±356.2          & 2841.7±356.4        & 2982.029±370.4 & B-IRL                & 2704.2±300.9          & 2433.9±285.2          & 2498.8±248.0          & 2488.4±226.6               \\
        BaseSVC              & 2465.2±257.4          & 2653.5±412.5          & 2608.6±447.9        & 2758.4±421.7 &  BaseSVC              & 2395.3±346.8          & 2131.7±376.1          & 2166.1±291.1          & 2170.2±299.2                  \\
        B-ICB-S              & 2637.6±168.5          & 2747.3±260.1          & 2805.3±398.6        & 2631.2±485.6 & B-ICB-S              & 2856.4±439.8          & 3281.5±162.1          & 3136.9±187.3          & 2980.8±350.3                     \\
        \hline
        \textbf{IBCB (ours)} & \textbf{2232.6±148.6} & \textbf{2071.0±264.4} & \textbf{2161.9±124.4} & \textbf{2192.0±110.0} & \textbf{IBCB (ours)} & \textbf{1596.7±172.3} & \textbf{1704.7±158.6} & \textbf{1584.4±124.1} & \textbf{1705.5±176.8} \\ 
        \hline
    \end{tabular}
    }

    \vspace{1em}

    \renewcommand{\arraystretch}{1.2}
    \scriptsize{\begin{tabular}{l|c|c|c|c||l|c|c|c|c}
        \hline
        \multirow{2}{*}{Algorithm} & \multicolumn{4}{c||}{Train Time (sec.)} & \multirow{2}{*}{Algorithm} & \multicolumn{4}{c}{Train Time (sec.)}\\
        \cline{2-5} \cline{7-10} ~ & std=0 & std=0.03 & std=0.07 & std=0.10 & ~ & std=0 & std=0.03 & std=0.07 & std=0.10 \\ 
        \hline \hline
        SBUCB                & /                    & /                    & /                    & /                                       & SBTS & /  & / & /  & /  \\
        \hline
        B-IRL                & 40.758±4.040         & 35.796±0.352         & 38.59±4.594          & 41.948±7.858         & B-IRL                & 35.664±0.208         & 37.07±2.135          & 39.162±6.970         & 35.662±0.081         \\
        BaseSVC              & 234.56±13.51       & 228.80±19.52       & 231.25±31.15        & 249.73±31.44   & BaseSVC              & 209.01±10.75       & 221.06±18.02       & 215.58±18.12       & 230.40±25.84      \\
        B-ICB-S              & 85.977±1.975         & 87.631±4.364         & 86.164±1.884         & 84.881±0.493        & B-ICB-S              & 86.726±1.870         & 87.447±1.977         & 86.796±1.808         & 85.974±1.826      \\
        \hline
        \textbf{IBCB (ours)} & \textbf{1.169±0.225} & \textbf{1.102±0.084} & \textbf{1.075±0.048} & \textbf{1.055±0.031} & \textbf{IBCB (ours)} & \textbf{0.798±0.020} & \textbf{0.786±0.012} & \textbf{0.791±0.013} & \textbf{0.849±0.116}  \\ 
        \hline
    \end{tabular}
    }
    
    
    
    \label{tab:fair:top:3:tot}
\end{table*}

\begin{table*}[!htbp]
    \centering
    \setlength{\abovecaptionskip}{5pt}
    \setlength{\belowcaptionskip}{-2pt}
    \caption{{ Fairness-aware} expert policy (SBUCB) parameters' estimation (\textbf{Upper Subtable}) and { Fairness-aware} expert policy (SBTS) parameters' estimation (\textbf{Lower Subtable})  on synthetic dataset. {Expert selects action from \textbf{top-$\bm 3$} actions with proportion of its learned probability distribution through all actions.} Expert policy produces same actions compared with itself, so Batch Test Log Fitness of expert policy is 1.}
    \renewcommand{\arraystretch}{1.2}
    \scriptsize{\begin{tabular}{l|c|c|c|c||l|c|c|c|c}
        \hline
        \multirow{2}{*}{Algorithm} & \multicolumn{4}{c||}{Batch Test Log Fitness} & \multirow{2}{*}{Algorithm} & \multicolumn{4}{c}{Batch Test Average Reward}\\
        \cline{2-5} \cline{7-10} ~ & std=0 & std=0.03 & std=0.07 & std=0.10 & ~ & std=0 & std=0.03 & std=0.07 & std=0.10\\ 
        \hline \hline
        SBUCB                & 1                    & 1                    & 1                    & 1                                        & SBUCB                & 0.538±0.012         & 0.529±0.030           & 0.528±0.031          & 0.516±0.031          \\
        \hline
        B-IRL                & 0.307±0.005          & 0.297±0.008         & 0.300±0.008            & 0.296±0.007         & B-IRL                & 0.486±0.018         & 0.472±0.039          & 0.472±0.041          & 0.457±0.042          \\
        BaseSVC              & 0.306±0.006          & 0.302±0.01          & 0.305±0.009          & 0.303±0.009          & BaseSVC              & 0.481±0.021         & 0.467±0.041          & 0.467±0.044          & 0.451±0.043          \\
        B-ICB-S              & 0.300±0.006            & 0.296±0.006         & 0.295±0.008          & 0.296±0.011         & B-ICB-S              & 0.500±0.004           & 0.484±0.025          & 0.482±0.029          & 0.475±0.026          \\
        \hline
        \textbf{IBCB (ours)} & \textbf{0.307±0.004} & \textbf{0.310±0.006} & \textbf{0.309±0.004} & \textbf{0.309±0.004} & \textbf{IBCB (ours)} & \textbf{0.558±0.010} & \textbf{0.543±0.029} & \textbf{0.545±0.027} & \textbf{0.535±0.029} \\ 
        \hline
    \end{tabular}
    }

    \vspace{1em}

    \renewcommand{\arraystretch}{1.2}
    \scriptsize{\begin{tabular}{l|c|c|c|c||l|c|c|c|c}
        \hline
        \multirow{2}{*}{Algorithm} & \multicolumn{4}{c||}{Batch Test Log Fitness} & \multirow{2}{*}{Algorithm} & \multicolumn{4}{c}{Batch Test Average Reward}\\
        \cline{2-5} \cline{7-10} ~ & std=0 & std=0.03 & std=0.07 & std=0.10 &  ~ & std=0 & std=0.03 & std=0.07 & std=0.10 \\ 
        \hline \hline
        SBTS                & 1                    & 1                    & 1                    & 1                                     &    SBTS                 & 0.548±0.028          & 0.566±0.030          & 0.560±0.025          & 0.556±0.024\\
        \hline
        B-IRL                & 0.302±0.006          & 0.309±0.003          & 0.305±0.002          & 0.307±0.002          & B-IRL                & 0.495±0.039          & 0.521±0.037          & 0.512±0.030          & 0.508±0.028          \\
        BaseSVC              & 0.308±0.007          & 0.311±0.005          & 0.312±0.006          & 0.312±0.004          & BaseSVC              & 0.492±0.039          & 0.518±0.039          & 0.510±0.032          & 0.507±0.031          \\
        B-ICB-S              & 0.292±0.011          & 0.281±0.008          & 0.287±0.004          & 0.292±0.009          & B-ICB-S              & 0.501±0.014          & 0.493±0.021          & 0.496±0.016          & 0.496±0.017          \\
        \hline
        \textbf{IBCB (ours)} & \textbf{0.321±0.004} & \textbf{0.319±0.004} & \textbf{0.322±0.003} & \textbf{0.318±0.004} & \textbf{IBCB (ours)} & \textbf{0.547±0.031} & \textbf{0.568±0.031} & \textbf{0.559±0.026} & \textbf{0.559±0.022}\\ 
        \hline
    \end{tabular}
    }
    
    
    \label{tab:fair:top:3:bt:bandit:tot}
\end{table*}

\begin{table*}[!htbp]
    \centering
    \setlength{\abovecaptionskip}{5pt}
    \setlength{\belowcaptionskip}{-2pt}
    \caption{{Cumulative Fairness Regret (CFR) comparison (\textbf{Upper Subtable}) and Train time comparison (\textbf{Lower Subtable}) on synthetic dataset. {Expert selects action from \textbf{top-$\bm 4$} actions with proportion of its learned probability distribution through all actions. Expert is defined as the fairest policy from the CFR metric, so its CFR is always 0. Fairness-aware experts (SBUCB \& SBTS) do not need to train since they produce the train logs (behavioral evolution history logs) for baselines and IBCB.}}}
    \renewcommand{\arraystretch}{1.2}
    \scriptsize{\begin{tabular}{l|c|c|c|c||l|c|c|c|c}
        \hline
        \multirow{2}{*}{Algorithm} & \multicolumn{4}{c||}{Cumulative Fairness Regret} & \multirow{2}{*}{Algorithm} & \multicolumn{4}{c}{Cumulative Fairness Regret}\\
        \cline{2-5} \cline{7-10} ~ & std=0 & std=0.03 & std=0.07 & std=0.10 &  ~ & std=0 & std=0.03 & std=0.07 & std=0.10\\ 
        \hline \hline
        SBUCB & 0 & 0 & 0 & 0 & SBTS & 0  & 0 & 0  & 0   \\
        \hline
        B-IRL                & 3063.9±258.1          & 3279.7±364.6          & 3214.4±381.0         & 3263.6±286.0          & B-IRL                & 2935.1±407.0          & 2812.5±254.6         & 2790.2±260.6        & 2794.1±319.6          \\
        BaseSVC              & \textbf{2893.6±240.7} & 3086.1±401.7          & 3042.2±399.8         & 3121.7±302.5          & BaseSVC              & 2752.0±379.1          & 2597.6±225.7         & 2617.2±273.2          & 2588.8±261.7          \\
        B-ICB-S              & 2998.0±153.3          & 3109.8±364.9            & 3123.4±287.4         & 3018.3±165.1  & B-ICB-S              & 3295.3±357.9          & 3092.2±240.2         & 3369.5±59.5           & 3093.9±388.0          \\
        \hline
        \textbf{IBCB (ours)} & 3130.6±194.2          & \textbf{2794.3±282.2} & \textbf{2966.9±205.4} & \textbf{2882.1±238.9} & \textbf{IBCB (ours)} & \textbf{2162.4±129.4} & \textbf{2117.3±43.8} & \textbf{1918.0±128.6} & \textbf{1990.1±232.6} \\ 
        \hline
    \end{tabular}
    }

    \vspace{1em}

    \renewcommand{\arraystretch}{1.2}
    \scriptsize{\begin{tabular}{l|c|c|c|c||l|c|c|c|c}
        \hline
        \multirow{2}{*}{Algorithm} & \multicolumn{4}{c||}{Train Time (sec.)} & \multirow{2}{*}{Algorithm} & \multicolumn{4}{c}{Train Time (sec.)} \\
        \cline{2-5} \cline{7-10} ~ & std=0 & std=0.03 & std=0.07 & std=0.10 & ~ & std=0 & std=0.03 & std=0.07 & std=0.10\\ 
        \hline \hline
        SBUCB  & /  & /  & /  & /  & SBTS & /  & / & /  & / \\
        \hline
        B-IRL                & 36.867±0.571 & 43.7±6.185              & 40.665±6.652            & 37.241±0.991           & B-IRL                & 36.697±0.373        & 41.568±4.424         & 39.308±5.713         & 36.270±0.706         \\
        BaseSVC              & 330.68±22.03        & 343.56±37.87 & 349.05±37.32 & 365.68±7.47 & BaseSVC              & 295.92±23.73      & 327.44±37.10       & 314.77±19.73       & 309.99±14.98      \\
        B-ICB-S              & 86.362±1.885          & 84.884±0.443            & 86.48±0.594             & 86.488±1.541           & B-ICB-S              & 88.003±2.241        & 86.189±2.22          & 86.318±1.266         & 87.87±1.607         \\
        \hline
        \textbf{IBCB (ours)} & \textbf{1.052±0.052}  & \textbf{1.016±0.037}    & \textbf{1.135±0.211}    & \textbf{1.034±0.032}   & \textbf{IBCB (ours)} & \textbf{0.950±0.175} & \textbf{0.737±0.013} & \textbf{0.838±0.148} & \textbf{0.820±0.112}\\ 
        \hline
    \end{tabular}
    }
    
    
    \label{tab:fair:top:4:tot}
\end{table*}

\begin{table*}[!htbp]
    \centering
    \setlength{\abovecaptionskip}{5pt}
    \setlength{\belowcaptionskip}{-2pt}
    \caption{{ Fairness-aware} expert policy (SBUCB) parameters' estimation (\textbf{Upper Subtable}) and { Fairness-aware} expert policy (SBTS) parameters' estimation (\textbf{Lower Subtable}) on synthetic dataset. {Expert selects action from \textbf{top-$\bm 4$} actions with proportion of its learned probability distribution through all actions.} Expert policy produces same actions compared with itself, so Batch Test Log Fitness of expert policy is 1.}
    \renewcommand{\arraystretch}{1.2}
    \scriptsize{\begin{tabular}{l|c|c|c|c||l|c|c|c|c}
        \hline
        \multirow{2}{*}{Algorithm} & \multicolumn{4}{c||}{Batch Test Log Fitness} & \multirow{2}{*}{Algorithm} & \multicolumn{4}{c}{Batch Test Average Reward}\\        \cline{2-5} \cline{7-10} ~ & std=0 & std=0.03 & std=0.07 & std=0.10 & ~ & std=0 & std=0.03 & std=0.07 & std=0.10 \\ 
        \hline \hline
        SBUCB                & 1                    & 1                    & 1                    & 1        &                                        SBUCB                & 0.490±0.01           & 0.480±0.029          & 0.478±0.029          & 0.471±0.028          \\ 
        \hline
        B-IRL                & \textbf{0.236±0.003} & 0.229±0.004          & 0.229±0.005          & 0.230±0.005           &        B-IRL                & 0.444±0.014          & 0.429±0.035          & 0.428±0.038          & 0.421±0.031          \\ 
        BaseSVC              & 0.235±0.003          & \textbf{0.232±0.007} & \textbf{0.232±0.005} & \textbf{0.233±0.006} &         BaseSVC              & 0.441±0.016          & 0.424±0.038          & 0.423±0.038          & 0.414±0.032          \\
        B-ICB-S              & 0.228±0.003          & 0.226±0.007          & 0.227±0.007          & 0.227±0.004 &          B-ICB-S              & 0.456±0.007          & 0.450±0.024          & 0.443±0.024          & 0.442±0.022               \\
        \hline
        \textbf{IBCB (ours)} & 0.227±0.003 & 0.231±0.007 & 0.230±0.003  & 0.230±0.004 &         \textbf{IBCB (ours)} & \textbf{0.523±0.005} & \textbf{0.499±0.024} & \textbf{0.502±0.024} & \textbf{0.495±0.024}  \\ 
        \hline
    \end{tabular}
    }

    \vspace{1em}

    \renewcommand{\arraystretch}{1.2}
    \scriptsize{\begin{tabular}{l|c|c|c|c||l|c|c|c|c}
        \hline
        \multirow{2}{*}{Algorithm} & \multicolumn{4}{c||}{Batch Test Log Fitness} & \multirow{2}{*}{Algorithm} & \multicolumn{4}{c}{Batch Test Average Reward}\\
        \cline{2-5} \cline{7-10} ~ & std=0 & std=0.03 & std=0.07 & std=0.10 & ~ & std=0 & std=0.03 & std=0.07 & std=0.10 \\ 
        \hline \hline
        SBTS                & 1                    & 1                    & 1                    & 1   & SBTS                 & 0.500±0.024          & 0.510±0.021          & 0.510±0.022          & 0.503±0.026                                      \\
        \hline
        B-IRL                & 0.234±0.006          & 0.234±0.003          & 0.235±0.002          & 0.233±0.002          & B-IRL                & 0.455±0.033          & 0.466±0.026          & 0.465±0.026          & 0.460±0.032          \\
        BaseSVC              & 0.235±0.004          & 0.238±0.005          & 0.235±0.003          & 0.235±0.004          & BaseSVC              & 0.453±0.034          & 0.466±0.027          & 0.464±0.028          & 0.458±0.032          \\
        B-ICB-S              & 0.226±0.006          & 0.229±0.007          & 0.222±0.007          & 0.228±0.009          & B-ICB-S              & 0.451±0.016          & 0.465±0.021          & 0.462±0.016          & 0.458±0.018          \\
        \hline
        \textbf{IBCB (ours)} & \textbf{0.239±0.004} & \textbf{0.241±0.002} & \textbf{0.242±0.004} & \textbf{0.241±0.003} & \textbf{IBCB (ours)} & \textbf{0.496±0.024} & \textbf{0.507±0.020} & \textbf{0.500±0.016} & \textbf{0.499±0.032} \\ 
        \hline
    \end{tabular}
    }
    
    
    \label{tab:fair:top:4:bt:bandit:tot}
\end{table*}

We also conducted fairness-aware expert experiments with different fairness settings on synthetic dataset. Specifically, in the original experiment, we set top-$k$ fairness with $k = 2$ (Table~\ref{tab:fair:top:2:tot}--Table~\ref{tab:fair:top:2:bt:bandit:tot}). We also make experiments with $k = 3, 4$ to discover the impact of changing fairness parameter (Table~\ref{tab:fair:top:3:tot}-Table~\ref{tab:fair:top:4:bt:bandit:tot}). It can be observed from the results that when $k$ increases, BC methods' training time significantly increase, indicating that BC methods tend to be confused because observed actions made by the expert are drawn from more possible candidates, i.e. the top-$k$ set. When $k$ increased, the BT-Fitness performance of all models decreased significantly, and the CFR also increased rapidly, which is related to the more random action selection of experts. However, IBCB can still quickly learn expert parameters and get the best performance in this situation, which also shows that IBCB still has a very high generalization ability in complex scenarios of fairness.
\subsubsection{Detailed Results of Contradictory Data Testing on Synthetic Dataset}\label{app:exp:details:dup}

\begin{table*}[!ht]
    \centering
    \caption{{Contradictory (duplicated) data (appeared at expert's OL-data behavioral evolution history logs) on synthetic dataset testing. $\mathrm{dup}$ denotes the quantity of duplicated episodes in OL-data phase. This table shows the OL-Fitness results.}}
    \renewcommand{\arraystretch}{1.2}
    \scriptsize{\begin{tabular}{l|c|c|c|c|c|c|c|c|c|c}
        \hline
        \multirow{2}{*}{Algorithm} & \multicolumn{9}{c}{Online Train Log Fitness} \\
        \cline{2-11} ~ & dup=0 & dup=1 & dup=2 & dup=3 & dup=4 & dup=5 & dup=6 & dup=7 & dup=8 & dup=9\\ 
        \hline \hline
        SBUCB & \textcolor[rgb]{0,0,0}{0.879±0.016} & \textcolor[rgb]{0,0,0}{0.880±0.015} & \textcolor[rgb]{0,0,0}{0.880±0.015} & \textcolor[rgb]{0,0,0}{0.879±0.015} & \textcolor[rgb]{0,0,0}{0.880±0.018} & \textcolor[rgb]{0,0,0}{0.879±0.015} & \textcolor[rgb]{0,0,0}{0.880±0.016} & \textcolor[rgb]{0,0,0}{0.879±0.015} & \textcolor[rgb]{0,0,0}{0.879±0.015} & \textcolor[rgb]{0,0,0}{0.878±0.017}\\
        \hline
        B-IRL & {0.808±0.039} & {0.808±0.037} & {0.808±0.038} & {0.806±0.038} & {0.807±0.038} & {0.807±0.037} & {0.809±0.037} & {0.810±0.037} & {0.809±0.037} & {0.808±0.037}  \\
        \hline
        \textbf{IBCB (Ours)} & \textbf{\textcolor[rgb]{0,0,0}{0.858±0.034}} & \textbf{\textcolor[rgb]{0,0,0}{0.860±0.021}} & \textbf{\textcolor[rgb]{0,0,0}{0.855±0.034}} & \textbf{\textcolor[rgb]{0,0,0}{0.865±0.021}} & \textbf{\textcolor[rgb]{0,0,0}{0.868±0.005}} & \textbf{\textcolor[rgb]{0,0,0}{0.874±0.003}} & \textbf{\textcolor[rgb]{0,0,0}{0.875±0.006}} & \textbf{\textcolor[rgb]{0,0,0}{0.848±0.038}} & \textbf{\textcolor[rgb]{0,0,0}{0.871±0.006}} & \textbf{\textcolor[rgb]{0,0,0}{0.867±0.008}}\\ 
        \hline
    \end{tabular}
    
    }
    
    \vspace{1em}
    
    \renewcommand{\arraystretch}{1.2}
    \scriptsize{\begin{tabular}{p{1.4cm}|c|c|c|c|c|c|c|c|c|c}
        \hline
        \multirow{2}{*}{Algorithm} & \multicolumn{9}{c}{Train Log Behaviors Fitness (BC)} \\
        \cline{2-11} ~ & dup=0 & dup=1 & dup=2 & dup=3 & dup=4 & dup=5 & dup=6 & dup=7 & dup=8 & dup=9 \\ 
        \hline \hline
        BaseSVC & 0.823±0.026 & 0.825±0.025 & 0.824±0.026 & 0.821±0.026 & 0.820±0.026 & 0.819±0.025 & 0.819±0.025 & 0.819±0.024 & 0.817±0.023 & 0.816±0.023  \\
        \hline
    \end{tabular}
    }

    \label{tab:ab:dup:OL}
\end{table*}

\begin{table*}[!ht]
    \centering
    \setlength{\abovecaptionskip}{3pt}
    \setlength{\belowcaptionskip}{-3pt}
    \caption{{Contradictory (duplicated) data (appeared at expert's OL-data behavioral evolution history logs) on synthetic dataset testing.  $\mathrm{dup}$ denotes the quantity of duplicated episodes in OL-data phase. This table shows the BT-Fitness results.}} 
    
    \renewcommand{\arraystretch}{1.2}
    \scriptsize{\begin{tabular}{l|c|c|c|c|c|c|c|c|c|c}
        \hline
        \multirow{2}{*}{Algorithm} & \multicolumn{10}{c}{Batch Test Log Fitness} \\
        \cline{2-11} ~ & dup=0 & dup=1 & dup=2 & dup=3 & dup=4 & dup=5 & dup=6 & dup=7 & dup=8 & dup=9 \\ 
        \hline \hline
        SBUCB & \textcolor[rgb]{0,0,0}{1} & \textcolor[rgb]{0,0,0}{1} & \textcolor[rgb]{0,0,0}{1} & \textcolor[rgb]{0,0,0}{1} & \textcolor[rgb]{0,0,0}{1} & \textcolor[rgb]{0,0,0}{1} & \textcolor[rgb]{0,0,0}{1} & \textcolor[rgb]{0,0,0}{1} & \textcolor[rgb]{0,0,0}{1} & \textcolor[rgb]{0,0,0}{1}\\
        \hline
        B-IRL & {0.820±0.013} & {0.820±0.012} & {0.820±0.013} & {0.820±0.013} & {0.819±0.014} & {0.818±0.014} & {0.819±0.014} & {0.818±0.011} & {0.818±0.011} & {0.819±0.014}  \\
        BaseSVC & 0.781±0.020 & 0.783±0.019 & 0.782±0.020 & 0.781±0.018 & 0.780±0.021 & 0.780±0.020 & 0.779±0.020 & 0.778±0.018 & 0.776±0.018 & 0.776±0.021  \\ 
        \hline
        \textbf{IBCB (Ours)} & \textbf{\textcolor[rgb]{0,0,0}{0.963±0.029}} & \textbf{\textcolor[rgb]{0,0,0}{0.964±0.015}} & \textbf{\textcolor[rgb]{0,0,0}{0.960±0.027}} & \textbf{\textcolor[rgb]{0,0,0}{0.967±0.017}} & \textbf{\textcolor[rgb]{0,0,0}{0.967±0.013}} & \textbf{\textcolor[rgb]{0,0,0}{0.972±0.015}} & \textbf{\textcolor[rgb]{0,0,0}{0.975±0.014}} & \textbf{\textcolor[rgb]{0,0,0}{0.953±0.034}} & \textbf{\textcolor[rgb]{0,0,0}{0.973±0.012}} & \textbf{\textcolor[rgb]{0,0,0}{0.973±0.013}}\\ 
        \hline
    \end{tabular}
    
    }
    \label{tab:ab:dup}
\end{table*}

Note that BaseSVC cannot obtain reward parameters, and thus were not included in OL-Fitness. We used `Train Log Behaviors Fitness' to evaluate whether BaseSVC suffers from contradictory data and make comparison with IBCB and other baselines. As shown in Table~\ref{tab:ab:dup:OL}, contradictory data (same context in different time periods) undermines BaseSVC's train log behaviors fitness, but IBCB stayed stable in OL-Fitness. That is, Contradictory data significantly affects BaseSVC's fitness but does not impact models that learn reward parameters like IBCB, highlighting IBCB's robustness in extreme conditions.

\label{app:exp:details:param}
\begin{table*}[!ht]
    \centering
    \setlength{\abovecaptionskip}{3pt}
    \setlength{\belowcaptionskip}{5pt}
    \caption{{Reward parameters' estimation (Left) and training time (Right) comparison on synthetic dataset (IBCB's parameter test). Experts do not need train since they produce train logs (behavioral evolution history logs) for baselines and IBCB. BaseSVC cannot obtain reward parameters, so it was excluded from OL-Fitness comparison. Original Expert is SBUCB.}}
    
    \renewcommand{\arraystretch}{1.2}
    \scriptsize{\begin{tabular}{l|c|c|c|c||l|c|c|c|c}
        \hline
        \multirow{2}{*}{Algorithm} & \multicolumn{4}{c||}{Online Train Log Fitness (Param Test)} & \multirow{2}{*}{Algorithm} & \multicolumn{4}{c}{Train Time(sec.) (Param Test)}  \\
        \cline{2-5} \cline{7-10} ~ & std=0 & std=0.03 & std=0.07 & std=0.10 & ~ & std=0 & std=0.03 & std=0.07 & std=0.10 \\ 
        \hline \hline
        SBUCB & \textcolor[rgb]{0,0,0}{0.883±0.007} & \textcolor[rgb]{0,0,0}{0.879±0.016} & \textcolor[rgb]{0,0,0}{0.878±0.018} & \textcolor[rgb]{0,0,0}{0.874±0.019} & SBUCB & \textcolor[rgb]{0,0,0}{/} & \textcolor[rgb]{0,0,0}{/} & \textcolor[rgb]{0,0,0}{/} & \textcolor[rgb]{0,0,0}{/}  \\ 
        \hline
        B-IRL & 0.827±0.012 & 0.808±0.039 & 0.806±0.037 & 0.802±0.036 & B-IRL & 36.997±0.803 & 37.670±1.328 & 39.910±3.419 & 37.424±0.563   \\ 
        BaseSVC & / & / & / & / & BaseSVC & 116.86±20.63 & 116.39±26.58 & 132.18±19.86 & 117.25±19.47  \\ 
        B-ICB-S & 0.805±0.020 & 0.771±0.031 & 0.780±0.034 & 0.771±0.024 & B-ICB-S & 86.365±0.827 & 87.599±4.489 & 86.85±1.162 & 85.713±0.676 \\
        \hline
        IBCB-$\alpha$-1	&	0.868±0.008 &	0.858±0.036 &	0.859±0.038	& 0.861±0.029 & IBCB-$\alpha$-1 & 0.815±0.045 &	0.999±0.128 & 0.866±0.137 & 1.016±0.109  \\
        IBCB-$\alpha$-0.6 &		\textbf{0.880±0.022} &	0.876±0.006 &	0.866±0.012 &	0.848±0.050 & IBCB-$\alpha$-0.6 & 0.907±0.109 & 0.971±0.092 & 0.904±0.106 &	0.867±0.121  \\
        IBCB-$\alpha$-0.8 &		\textbf{0.880±0.022} &	0.876±0.006 &	0.866±0.012 &	0.848±0.050 & IBCB-$\alpha$-0.8	&	0.958±0.134 &	0.880±0.097 &	0.868±0.086 &	0.961±0.119  \\
        IBCB-$\alpha$-1.2 &		0.867±0.005 &	0.870±0.007 &	0.869±0.010 &	0.862±0.009 & IBCB-$\alpha$-1.2 &		0.725±0.037 &	0.776±0.036 &	0.730±0.025 &	0.772±0.041  \\
        IBCB-$\alpha$-1.5 &		0.879±0.012 &	\textbf{0.880±0.01}3 &	\textbf{0.870±0.018 }&	\textbf{0.874±0.014} & IBCB-$\alpha$-1.5 &		\textbf{0.462±0.052} &	\textbf{0.467±0.054} &	\textbf{0.469±0.044} &	\textbf{0.468±0.046}  \\
        IBCB-$\alpha$-2	 &	0.865±0.007 &	0.865±0.012 &	0.864±0.016 &	0.861±0.008 & IBCB-$\alpha$-2 &		0.574±0.007 &	0.582±0.017 &	0.557±0.052 &	0.538±0.070   \\
        \hline
    \end{tabular}
    
    }
    \label{tab:ab:param:real}
\end{table*}

\begin{table*}[!ht]
    \centering
    \setlength{\abovecaptionskip}{3pt}
    \setlength{\belowcaptionskip}{5pt}
    \caption{{Expert policy parameters' estimation on synthetic dataset (IBCB's parameter test). Expert policy produces same actions compared with itself, so BT-Fitness of expert policy is 1. Original Expert is SBUCB.}}
    
    \renewcommand{\arraystretch}{1.2}
    \scriptsize{\begin{tabular}{l|c|c|c|c||l|c|c|c|c}
        \hline
        \multirow{2}{*}{Algorithm} & \multicolumn{4}{c||}{Batch Test Log Fitness (Param Test)} & \multirow{2}{*}{Algorithm} & \multicolumn{4}{c}{Batch Test Average Reward (Param Test)} \\
        \cline{2-5} \cline{7-10} ~ & std=0 & std=0.03 & std=0.07 & std=0.10 & ~ & std=0 & std=0.03 & std=0.07 & std=0.10\\ 
        \hline \hline
        SBUCB & \textcolor[rgb]{0,0,0}{1} & \textcolor[rgb]{0,0,0}{1} & \textcolor[rgb]{0,0,0}{1} & \textcolor[rgb]{0,0,0}{1} & SBUCB & \textcolor[rgb]{0,0,0}{0.641±0.010} & \textcolor[rgb]{0,0,0}{0.629±0.028} & \textcolor[rgb]{0,0,0}{0.630±0.024} & \textcolor[rgb]{0,0,0}{0.622±0.024}    \\ 
        \hline
        B-IRL & 0.828±0.005 & 0.820±0.013 & 0.820±0.014 & 0.818±0.016 & B-IRL & 0.588±0.015 & {0.571±0.037} & 0.572±0.034 & {0.561±0.034}   \\ 
        BaseSVC & 0.791±0.008 & 0.781±0.020 & 0.781±0.020 & 0.777±0.002 & BaseSVC & 0.577±0.017 & 0.559±0.042 & 0.559±0.037 & 0.547±0.037  \\ 
        B-ICB-S & 0.800±0.032 & 0.767±0.055 & 0.780±0.045 & 0.776±0.022 & B-ICB-S & 0.581±0.006 & 0.553±0.024 & 0.560±0.025 & 0.552±0.025   \\
        \hline
        IBCB-$\alpha$-1 &		0.970±0.012 &	0.960±0.036 &	0.959±0.034 &	\textbf{0.969±0.020} & IBCB-$\alpha$-1	&	0.647±0.012 &	0.638±0.026 &	0.639±0.019 &	0.628±0.022 \\
        IBCB-$\alpha$-0.6	&	0.964±0.022 &	0.966±0.017 &	0.950±0.015 &	0.933±0.047 & IBCB-$\alpha$-0.6 &		0.644±0.013 &	0.634±0.032 &	0.639±0.03 &	\textbf{0.637±0.018 } \\
        IBCB-$\alpha$-0.8	&	0.969±0.017 &	0.954±0.032 &	0.937±0.042 &	0.930±0.041 & IBCB-$\alpha$-0.8 &		0.647±0.012 &	\textbf{0.640±0.025} &	\textbf{0.644±0.02} &	0.635±0.020  \\
        IBCB-$\alpha$-1.2	&	\textbf{0.972±0.006 }&	0.971±0.012 &	\textbf{0.970±0.005} &	0.967±0.013 & IBCB-$\alpha$-1.2	&	0.647±0.012 &	0.632±0.032 &	0.633±0.029 &	0.625±0.030 \\
        IBCB-$\alpha$-1.5	&	0.959±0.011 &	0.960±0.012 &	0.943±0.015 &	0.956±0.020 & IBCB-$\alpha$-1.5	&	\textbf{0.650±0.010} &	0.637±0.029 &	0.643±0.020 &	0.631±0.025 \\
        IBCB-$\alpha$-2	&	0.970±0.006 &	\textbf{0.973±0.017 }&	0.962±0.018 &	0.960±0.021 & IBCB-$\alpha$-2	&	0.647±0.011 &	0.633±0.031 &	0.638±0.020 &	0.629±0.025  \\
        \hline
    \end{tabular}
    
    }
    \label{tab:ab:param:bt:bandit:SBUCB:tot}
\end{table*}
\subsubsection{IBCB's Parameter Test on Synthetic Dataset}
\label{app:exp:details:ab:param}
`IBCB-$\alpha$-\textit{k}' denotes IBCB's hyper parameter (in Eq.\eqref{eq:inverse:opt:constriant}) is \textit{k}. As shown in Table~\ref{tab:ab:param:real} and Table~\ref{tab:ab:param:bt:bandit:SBUCB:tot}, IBCB's hyper parameter $\alpha$ will influence several indicators, especially the train time, due to the OSQP tolerance level will be relaxed if $\alpha$ is above 1.5. So for IBCB of large $\alpha$ with more relaxed tolerance, $i.e.,$ \textit{lower accuracy}, training will be faster. Meanwhile, the performance of expert policy's parameter estimation is similar according to BT-Fitness and BT-AR. By examining the parameters learned by IBCB with different $\alpha$ values, we found that after normalization (by dividing each parameter's $l_2$-norm), the sum of absolute differences in every dimension of the parameters from each pair of two distinct parameters is tightly small. This suggests that one of these parameters can be a result of scaling up or down from another. During BT-data testing, items are recommended based on the parameter under BCB setting without exploration. Therefore, scaling up or down does not influence the recommended item since $\operatornamewithlimits{arg\,max}_{I \in \mathcal{I}} \langle \hat{\bm\theta}, \bm s_I \rangle$ is only determined by the item's content $\bm s_I$ when $\hat{\bm\theta}$ is fixed with a scaling factor. On the synthetic dataset, the item's content $\bm s_I$ is very easy to separate (interval difference of each dimension of $\bm s_I$ is very obvious through the setting in Section~\ref{exp:dataset:syn}), so two distinct parameters with very small sum of absolute differences may share similar BT-Fitness and BT-AR.

\subsubsection{Ablation Study on ML-100K Dataset}
\label{app:exp:details:ab:ML}
\begin{table}[!ht]
    \centering
    \setlength{\abovecaptionskip}{3pt}
    \setlength{\belowcaptionskip}{5pt}
    \caption{{Reward \& policy parameters' estimation on ML-100K dataset (using first half of expert's behavioral evolution history logs). Expert policy produces same actions compared with itself, so BT-Fitness of expert policy is 1. BaseSVC cannot obtain reward parameters, so it was excluded from OL-Fitness comparison.}}
    \renewcommand{\arraystretch}{1.2}
        \scriptsize{\begin{tabular}{l|c|c|c}
        \hline
        Algorithm & OL-Fitness & BT-Fitness & BT-AR \\
        \hline \hline
        SBUCB & \textcolor[rgb]{0,0,0}{0.891±0.006} & \textcolor[rgb]{0,0,0}{1} & \textcolor[rgb]{0,0,0}{0.193±0.005} \\ 
        \hline
        B-IRL & {0.789±0.017} & {0.784±0.022} & {0.189±0.004} \\ 
        BaseSVC & / & 0.701±0.022 & 0.186±0.004 \\ 
        B-ICB-S & 0.399±0.110 & 0.325±0.123 & 0.140±0.025 \\ 
        \hline
        \textbf{IBCB (Ours)} & \textbf{\textcolor[rgb]{0,0,0}{0.861±0.002}} & \textbf{\textcolor[rgb]{0,0,0}{0.829±0.018}} & \textbf{\textcolor[rgb]{0,0,0}{0.189±0.006}} \\ 
        \hline
    \end{tabular}   
    }
    \label{tab:ML:ab:ce}
\end{table}

\begin{table}[!ht]
    \centering
    \setlength{\abovecaptionskip}{3pt}
    \setlength{\belowcaptionskip}{5pt}
    \caption{{IBCB's parameter test on ML-100K dataset. Experts do not need to train since they produce train logs (behavioral evolution history logs) for baselines and IBCB.}}
    \renewcommand{\arraystretch}{1.2}
    \scriptsize{\begin{tabular}{l|c|c|c|c}
        \hline
        Algorithm & OL-Fitness & BT-Fitness & BT-AR & Train Time (sec.) \\
        \hline \hline
        SBUCB & \textcolor[rgb]{0,0,0}{0.891±0.006} & \textcolor[rgb]{0,0,0}{1} & \textcolor[rgb]{0,0,0}{0.193±0.005} & \textcolor[rgb]{0,0,0}{/}\\
        \hline
        B-IRL & 0.830±0.010 & 0.864±0.008 & {0.189±0.005} & 43.669±1.282 \\ 
        BaseSVC & / & 0.804±0.012 & {0.189±0.005} & 162.767±12.135 \\ 
        B-ICB-S & 0.746±0.032 & 0.767±0.028 & 0.187±0.007 & 147.474±4.633 \\ 
        \hline
        IBCB-$\alpha$-1	&	\textbf{0.862±0.023} &	0.925±0.019 &	0.192±0.005 &	4.036±0.111 \\
        IBCB-$\alpha$-0.6 &		0.775±0.007 &	0.861±0.016 &	0.192±0.005 &	3.648±0.139 \\
        IBCB-$\alpha$-0.8	&	0.860±0.029 &	0.921±0.021 &	0.191±0.006 &	4.026±0.194 \\
        IBCB-$\alpha$-1.2	&	0.861±0.013 &	\textbf{0.929±0.007} &	0.191±0.006 &	4.210±0.121 \\
        IBCB-$\alpha$-1.5	&	0.858±0.014 &	0.855±0.012 &	\textbf{0.194±0.005} &	\textbf{2.501±0.075} \\
        IBCB-$\alpha$-2	&	0.845±0.017 &	0.855±0.012 &	\textbf{0.194±0.005} &	2.663±0.221 \\
        \hline
    \end{tabular}
    }
    \label{tab:ML:ab:param}
\end{table}
For ML-100K dataset, Table~\ref{tab:ML:ab:ce} and Table~\ref{tab:ML:ab:param}  reports the result of different ablation studies. Results are similar to those on synthetic dataset in most indicators: IBCB can learn from novice expert's evolution history with better performance compared with other baselines. Note that mean and variance of train time for IBCB-$\alpha$-1.2 is larger than any other $\alpha$ settings. From the OSQP train record, we analysed that this happened due to OSQP failed to convergent for some of the OL-data train logs, so for large $\alpha$, it is recommended to use more relaxed tolerance level in OSQP setting. Note that B-ICB-S's performance significantly decline when using first half of the train logs in Table~\ref{tab:ML:ab:ce}, this is because when in first half of the real-world train logs, \textit{novice expert} is still exploring with large steps in a batched manner rather than updating after each step or staying stable with little exploration, which B-ICB-S assumed.

\subsubsection{Extra Ablation Study Experiments of Batch Test period on ML-100K Dataset}
\label{app:exp:details:ab:ML:extra}
Furthermore, we conducted some extra ablation study experiments on ML-100K to test the real-world task's performance (e.g. train time, average reward in Batch Test period) improvements against other baselines. 
\begin{table}[!ht]
    \centering
    \setlength{\abovecaptionskip}{3pt}
    \setlength{\belowcaptionskip}{5pt}
    \caption{{
    Comparison made between original algorithms and baselines using average method when training on ML-100K dataset. \textbf{Upper Subtable} is the performance made by original ones, which use \textbf{whole} data to train. \textbf{Lower Subtable} is the performance made by algorithms using average method (only baselines, excluding SBUCB and IBCB), which use every episode data training distinct models, and final model uses \textbf{simple average} method to average these models' parameters.}}
    
    \renewcommand{\arraystretch}{1.2}
    \scriptsize{\begin{tabular}{l|c|c|c|c}
        \hline
        Algorithm & OL-Fitness & BT-Fitness & BT-AR & Train Time (sec.) \\
        \hline \hline
        SBUCB & \textcolor[rgb]{0,0,0}{0.891±0.006} & 1 & \textcolor[rgb]{0,0,0}{0.193±0.005} & {/} \\ 
        \hline
        B-IRL & {0.830±0.010} & 0.864±0.008 & {0.189±0.005} & {43.669±1.282}\\ 
        BaseSVC & / & 0.804±0.012 & {0.189±0.005} & 162.767±12.135 \\ 
        B-ICB-S & 0.746±0.032 & 0.767±0.028 & {0.187±0.007} & 147.474±4.633 \\
        \hline
        \textbf{IBCB (Ours)} & \textbf{\textcolor[rgb]{0,0,0}{0.862±0.023}} & \textbf{0.925±0.019} & \textbf{0.192±0.005} & \textbf{\textcolor[rgb]{0,0,0}{4.036±0.111}} \\ 
        \hline
    \end{tabular}
    }

    \vspace{1em}

    
    \renewcommand{\arraystretch}{1.2}
    \scriptsize{\begin{tabular}{l|c|c|c|c}
        \hline
        Algorithm & OL-Fitness & BT-Fitness & BT-AR & Train Time (sec.) \\
        \hline \hline
        SBUCB & \textcolor[rgb]{0,0,0}{0.891±0.006} & 1 & \textcolor[rgb]{0,0,0}{0.193±0.005} & {/}\\ 
        \hline
        B-IRL & {0.835±0.007} & 0.868±0.008 & {0.190±0.005}  & {85.352±9.192} \\ 
        BaseSVC & / & 0.748±0.009 & {0.189±0.004}  & 20.619±0.377 \\ 
        B-ICB-S & 0.336±0.076 & 0.262±0.064 & {0.129±0.019} & 214.288±2.558 \\
        \hline
        \textbf{IBCB (Ours)} & \textbf{\textcolor[rgb]{0,0,0}{0.862±0.023}} & \textbf{0.925±0.019} & \textbf{0.192±0.005} & \textbf{\textcolor[rgb]{0,0,0}{4.036±0.111}} \\ 
        \hline
    \end{tabular}
    }
    \label{tab:ML:ex:avg}
\end{table}

\begin{table}[!ht]
    \centering
    \setlength{\abovecaptionskip}{0pt}
    \setlength{\belowcaptionskip}{0pt}
    \caption{{
    Comparison made between larger batch size $B$ ($B=2000$) and smaller batch size $B$ ($B=500$)  when training \& testing on ML-100K dataset. \textbf{Upper Subtable} is the experiment results obtained when $B=2000$. \textbf{Lower Subtable} is the experiment results obtained when $B=500$.}}
        
    
    \renewcommand{\arraystretch}{1.2}
    \scriptsize{\begin{tabular}{l|c|c|c|c}
        \hline
        Algorithm & OL-Fitness & BT-Fitness & BT-AR & Train Time (sec.) \\
        \hline \hline
        SBUCB & \textcolor[rgb]{0,0,0}{0.896±0.001} & 1 & \textcolor[rgb]{0,0,0}{0.194±0.005} & {/} \\ 
        \hline
        B-IRL & {0.816±0.023} & {0.866±0.014} & {0.192±0.005} & {52.485±1.241} \\ 
        BaseSVC & / & 0.807±0.014 & {0.191±0.004} & 230.26±12.606 \\ 
        B-ICB-S & 0.685±0.028 & 0.707±0.052 & {0.188±0.006} & 144.697±7.972 \\
        \hline
        \textbf{IBCB (Ours)} & \textbf{\textcolor[rgb]{0,0,0}{0.861±0.010}} & \textbf{0.886±0.014} & \textbf{0.192±0.004} & \textbf{\textcolor[rgb]{0,0,0}{2.907±0.261}} \\ 
        \hline
    \end{tabular}
    
    }

    \vspace{1em}

        
    \centering
    
    \renewcommand{\arraystretch}{1.2}
    \scriptsize{\begin{tabular}{l|c|c|c|c}
        \hline
        Algorithm & OL-Fitness & BT-Fitness & BT-AR & Train Time (sec.) \\
        \hline \hline
         SBUCB & \textcolor[rgb]{0,0,0}{0.890±0.012} & 1 & \textcolor[rgb]{0,0,0}{0.193±0.006} & {/}\\ 
        \hline
        B-IRL & \textbf{0.845±0.007} & \textbf{0.877±0.009} & {0.190±0.006} & {65.646±1.378} \\ 
        BaseSVC & / & 0.815±0.006 & {0.189±0.005}  & 266.538±12.153 \\ 
        B-ICB-S & 0.716±0.039 & 0.723±0.049 & {0.187±0.007} & 138.616±8.982 \\
        \hline
        \textbf{IBCB (Ours)} & 0.835±0.023 & \textcolor[rgb]{0,0,0}{0.833±0.019} & \textbf{0.194±0.006} & \textbf{\textcolor[rgb]{0,0,0}{4.296±0.128}} \\ 
        \hline
    \end{tabular}
    
    }
    
    \label{tab:ML:ex:btsz}
\end{table}

Firstly, since we assume that training period can be described as a period containing $N$ episodes' training data, we want to know how this $N$'s setting influence the performance of baselines. Specifically, we divided the total $N$ episodes to individually train corresponding IRL or BC models. Finally, we averaged the parameters of the $N$ models obtained from each episode's training to form the final test model. In this setup, the training time equates to the cumulative time for training $N$ models (disregarding multi-threading, assuming that each time we collect data from an episode, we train a model sequentially). The experiment was conducted using the ML-100K dataset, as we aimed to investigate the practical feasibility. Detailed experimental results can be found in Table~\ref{tab:ML:ex:avg}. From the experimental results, it is evident that training IRL or BC models separately for each episode does yield certain differences in comparison to the unified training approach. The B-IRL model demonstrates a slight improvement in performance (e.g. OL-Fitness), albeit with a significant increase in training time. This could be attributed to the necessity of restarting training each time due to the inability to iteratively utilize previously acquired parameters when new data is collected. Of particular interest is the significant decline in performance for B-ICB. This discrepancy is likely linked to the inconsistency in retaining the original expert model's (i.e. SBUCB) parameters within each episode, which contradicts B-ICB's assumption of evolving data within every single step. Additionally, due to the necessity for repeated restarts during training, the overall training time becomes extended. As for BaseSVC, the reduced training sample size in each episode leads to notably shorter training time for individual runs (attributed to its near-quadratic time complexity). While this reduces overall training time, the efficacy is comparatively diminished compared to the direct training using the entire dataset. Ultimately, even in comparison with the models subjected to parameter averaging, the performance of IBCB remains superior overall. This underscores the effectiveness of using episodes as evolutionary segments in IBCB's training process in real-world datasets.

Secondly, we wanted to work out the influence of batch size $B$'s setting. Throughout this ablation study experiment, we controlled the total number of training and testing samples and fixed the context, ensuring that the product of the total number of episodes $N$ and the batch size $B$ within each episode remains constant. Based on this foundation, we adjusted the batch size $B$. Specifically, the original values were $N = 20$ and $B = 1000$. Extra experiments included scenarios with larger $B$, denoted as $N = 10$ and $B= 2000$ , with corresponding results displayed in upper subtable of Table~\ref{tab:ML:ex:btsz}. Likewise, there were scenarios with smaller $B$, represented as $N = 40$ and $B = 500$, with results exhibited in lower subtable of Table~\ref{tab:ML:ex:btsz}. From the results of these additional experiments and comparisons with the original experiments, we deduced that, while controlling the total number of training and testing samples and context, a larger $B$ yields shorter overall training time and better fitness on both environment and expert algorithm's parameters. Therefore, the fitting to the environment and expert is reduced due to the increased value of $N$, implying that more data for training comprises novice expert's evolutionary history, hence fostering more exploration. With reduced $N$ and increased $B$, the frequency of evolution decreases, making adaptation to the original training data easier, thereby shortening training time. We found that the smaller $B$, the larger the total number of episodes $N$, but the extent of subsequent episode exploration gradually becomes smaller, so IBCB may tend to overfit. On the contrary, when the $B$ is larger, the smaller the total number of episodes $N$, so the environment will encourage more exploration, and IBCB can fit better environment and expert algorithm's parameters. In such cases, IBCB shows preciser performance against other baselines under larger $B$. Appropriately crafting exploration items can result in higher rewards. 

\section{Conclusion}
This paper aims to address the problem of efficiently learning behaviours from the expert's evolution history. Specifically, we propose an inverse batched contextual bandit model called IBCB. The proposed IBCB gives a unified framework for both deterministic and randomized bandit policies, and solves the problem of learning from evolution history with inaccessible rewards through a simple quadratic programming problem. Experimental results demonstrate the effectiveness and efficient training of IBCB on both synthetic and real-world scenarios. Besides, IBCB can also be generalized to out-of-distribution and contradictory data scenarios with great robustness. When it comes to limitations, IBCB needs to be further explored in experiments related to fairness, such as more random action selection methods, fairness assumptions based on specific distributions and other related settings. In addition, this article mainly discusses the UCB-like gambling machine framework. In the future work, we will focus on better random optimization of IBCB.
\bibliographystyle{IEEEtran}
\bibliography{bibtex/bib/IBCB_bib}

\begin{thebibliography}{10}
\providecommand{\url}[1]{#1}
\csname url@samestyle\endcsname
\providecommand{\newblock}{\relax}
\providecommand{\bibinfo}[2]{#2}
\providecommand{\BIBentrySTDinterwordspacing}{\spaceskip=0pt\relax}
\providecommand{\BIBentryALTinterwordstretchfactor}{4}
\providecommand{\BIBentryALTinterwordspacing}{\spaceskip=\fontdimen2\font plus
\BIBentryALTinterwordstretchfactor\fontdimen3\font minus \fontdimen4\font\relax}
\providecommand{\BIBforeignlanguage}[2]{{%
\expandafter\ifx\csname l@#1\endcsname\relax
\typeout{** WARNING: IEEEtran.bst: No hyphenation pattern has been}%
\typeout{** loaded for the language `#1'. Using the pattern for}%
\typeout{** the default language instead.}%
\else
\language=\csname l@#1\endcsname
\fi
#2}}
\providecommand{\BIBdecl}{\relax}
\BIBdecl

\bibitem{Bain1995Framework}
M.~Bain and C.~Sammut, ``A framework for behavioural cloning.'' in \emph{Machine Intelligence 15}, 1995, pp. 103--129.

\bibitem{Russell1998Learning}
S.~Russell, ``Learning agents for uncertain environments,'' in \emph{Proceedings of the eleventh annual conference on Computational learning theory}, 1998, pp. 101--103.

\bibitem{Zheng2022Imitation}
B.~Zheng, S.~Verma, J.~Zhou, I.~W. Tsang, and F.~Chen, ``Imitation learning: Progress, taxonomies and challenges,'' \emph{IEEE Transactions on Neural Networks and Learning Systems}, pp. 1--16, 2022.

\bibitem{Abbeel2004Apprenticeship}
P.~Abbeel and A.~Y. Ng, ``Apprenticeship learning via inverse reinforcement learning,'' in \emph{Proceedings of the twenty-first international conference on Machine learning}, 2004, p.~1.

\bibitem{Hussein2017Imitation}
A.~Hussein, M.~M. Gaber, E.~Elyan, and C.~Jayne, ``Imitation learning: A survey of learning methods,'' \emph{ACM Computing Surveys (CSUR)}, vol.~50, no.~2, pp. 1--35, 2017.

\bibitem{Zhang2021Counterfactual}
X.~Zhang, H.~Jia, H.~Su, W.~Wang, J.~Xu, and J.~Wen, ``Counterfactual reward modification for streaming recommendation with delayed feedback,'' in \emph{Proceedings of the 44th International {ACM} {SIGIR} Conference on Research and Development in Information Retrieval}, 2021, pp. 41--50.

\bibitem{Zhang2022Counteracting}
X.~Zhang, S.~Dai, J.~Xu, Z.~Dong, Q.~Dai, and J.-R. Wen, ``Counteracting user attention bias in music streaming recommendation via reward modification,'' in \emph{Proceedings of the 28th ACM SIGKDD Conference on Knowledge Discovery and Data Mining}, 2022, pp. 2504--2514.

\bibitem{Chandramouli2011Streamrec}
B.~Chandramouli, J.~J. Levandoski, A.~Eldawy, and M.~F. Mokbel, ``Streamrec: a real-time recommender system,'' in \emph{Proceedings of the 2011 ACM SIGMOD International Conference on Management of data}, 2011, pp. 1243--1246.

\bibitem{Chang2017Streaming}
S.~Chang, Y.~Zhang, J.~Tang, D.~Yin, Y.~Chang, M.~A. Hasegawa-Johnson, and T.~S. Huang, ``Streaming recommender systems,'' in \emph{Proceedings of the 26th international conference on world wide web}, 2017, pp. 381--389.

\bibitem{Jakomin2020Simultaneous}
M.~Jakomin, Z.~Bosni{\'c}, and T.~Curk, ``Simultaneous incremental matrix factorization for streaming recommender systems,'' \emph{Expert Systems with Applications}, vol. 160, p. 113685, 2020.

\bibitem{wang2021fairness}
L.~Wang, Y.~Bai, W.~Sun, and T.~Joachims, ``Fairness of exposure in stochastic bandits,'' in \emph{International Conference on Machine Learning}.\hskip 1em plus 0.5em minus 0.4em\relax PMLR, 2021, pp. 10\,686--10\,696.

\bibitem{gillen2018online}
S.~Gillen, C.~Jung, M.~Kearns, and A.~Roth, ``Online learning with an unknown fairness metric,'' \emph{Advances in neural information processing systems}, vol.~31, 2018.

\bibitem{Ramachandran2007Bayesian}
D.~Ramachandran and E.~Amir, ``Bayesian inverse reinforcement learning.'' in \emph{IJCAI}, vol.~7, 2007, pp. 2586--2591.

\bibitem{Choi2011MAP}
J.~Choi and K.-E. Kim, ``Map inference for bayesian inverse reinforcement learning,'' \emph{Advances in neural information processing systems}, vol.~24, 2011.

\bibitem{Huyuk2022Inverse}
A.~H{\"u}y{\"u}k, D.~Jarrett, and M.~van~der Schaar, ``Inverse contextual bandits: Learning how behavior evolves over time,'' in \emph{International Conference on Machine Learning}.\hskip 1em plus 0.5em minus 0.4em\relax PMLR, 2022, pp. 9506--9524.

\bibitem{Han2020Sequential}
Y.~Han, Z.~Zhou, Z.~Zhou, J.~H. Blanchet, P.~W. Glynn, and Y.~Ye, ``Sequential batch learning in finite-action linear contextual bandits,'' \emph{CoRR}, vol. abs/2004.06321, 2020.

\bibitem{Dimakopoulou2019Balanced}
M.~Dimakopoulou, Z.~Zhou, S.~Athey, and G.~Imbens, ``Balanced linear contextual bandits,'' in \emph{Proceedings of the 33rd {AAAI} Conference on Artificial Intelligence}, 2019, pp. 3445--3453.

\bibitem{Li2010Contextual}
L.~Li, W.~Chu, J.~Langford, and R.~E. Schapire, ``A contextual-bandit approach to personalized news article recommendation,'' in \emph{Proceedings of the 19th International Conference on World Wide Web}, 2010, pp. 661--670.

\bibitem{Lan2016Contextual}
A.~S. Lan and R.~G. Baraniuk, ``A contextual bandits framework for personalized learning action selection,'' in \emph{Proceedings of the 9th International Conference on Educational Data Mining}, 2016, pp. 424--429.

\bibitem{Yang2021Impact}
J.~Yang, W.~Hu, J.~D. Lee, and S.~S. Du, ``Impact of representation learning in linear bandits,'' in \emph{Proceedings of the 9th International Conference on Learning Representations}, 2021.

\bibitem{Ren2020Dynamic}
Z.~Ren and Z.~Zhou, ``Dynamic batch learning in high-dimensional sparse linear contextual bandits,'' \emph{arXiv preprint arXiv:2008.11918}, 2020.

\bibitem{Gu2021Batched}
Q.~Gu, A.~Karbasi, K.~Khosravi, V.~Mirrokni, and D.~Zhou, ``Batched neural bandits,'' \emph{arXiv preprint arXiv:2102.13028}, 2021.

\bibitem{Arora2021Survey}
S.~Arora and P.~Doshi, ``A survey of inverse reinforcement learning: Challenges, methods and progress,'' \emph{Artificial Intelligence}, vol. 297, p. 103500, 2021.

\bibitem{Abbeel2010Autonomous}
P.~Abbeel, A.~Coates, and A.~Y. Ng, ``Autonomous helicopter aerobatics through apprenticeship learning,'' \emph{The International Journal of Robotics Research}, vol.~29, no.~13, pp. 1608--1639, 2010.

\bibitem{Osa2017Online}
T.~Osa, N.~Sugita, and M.~Mitsuishi, ``Online trajectory planning and force control for automation of surgical tasks,'' \emph{IEEE Transactions on Automation Science and Engineering}, vol.~15, no.~2, pp. 675--691, 2017.

\bibitem{Liu2020Imitation}
E.~Liu, M.~Hashemi, K.~Swersky, P.~Ranganathan, and J.~Ahn, ``An imitation learning approach for cache replacement,'' in \emph{International Conference on Machine Learning}.\hskip 1em plus 0.5em minus 0.4em\relax PMLR, 2020, pp. 6237--6247.

\bibitem{balakrishna2020policy}
A.~Balakrishna, B.~Thananjeyan, J.~Lee, F.~Li, A.~Zahed, J.~E. Gonzalez, and K.~Goldberg, ``On-policy robot imitation learning from a converging supervisor,'' in \emph{Conference on Robot Learning}.\hskip 1em plus 0.5em minus 0.4em\relax PMLR, 2020, pp. 24--41.

\bibitem{Ziebart2008Maximum}
B.~D. Ziebart, A.~L. Maas, J.~A. Bagnell, A.~K. Dey \emph{et~al.}, ``Maximum entropy inverse reinforcement learning.'' in \emph{Aaai}, vol.~8.\hskip 1em plus 0.5em minus 0.4em\relax Chicago, IL, USA, 2008, pp. 1433--1438.

\bibitem{Fu2018Learning}
J.~Fu, K.~Luo, and S.~Levine, ``Learning robust rewards with adverserial inverse reinforcement learning,'' in \emph{Proceedings of the 6th International Conference on Learning Representations}, 2018.

\bibitem{Qureshi2019Adversarial}
A.~H. Qureshi, B.~Boots, and M.~C. Yip, ``Adversarial imitation via variational inverse reinforcement learning,'' in \emph{Proceedings of the 7th International Conference on Learning Representations}, 2019.

\bibitem{Ho2016Generative}
J.~Ho and S.~Ermon, ``Generative adversarial imitation learning,'' \emph{Advances in Neural Information Processing Systems}, vol.~29, 2016.

\bibitem{Chen2021Generative}
X.~Chen, L.~Yao, A.~Sun, X.~Wang, X.~Xu, and L.~Zhu, ``Generative inverse deep reinforcement learning for online recommendation,'' in \emph{Proceedings of the 30th ACM International Conference on Information \& Knowledge Management}, 2021, pp. 201--210.

\bibitem{Brown2019Deep}
D.~S. Brown and S.~Niekum, ``Deep bayesian reward learning from preferences,'' \emph{Workshop on Safety and Robustness in Decision-Making at the 33rd Conference on Neural Information Processing Systems (NeurIPS) 2019}, 2019.

\bibitem{Prudencio2023Survey}
R.~F. Prudencio, M.~R. Maximo, and E.~L. Colombini, ``A survey on offline reinforcement learning: Taxonomy, review, and open problems,'' \emph{IEEE Transactions on Neural Networks and Learning Systems}, 2023.

\bibitem{Levine2020Offline}
S.~Levine, A.~Kumar, G.~Tucker, and J.~Fu, ``Offline reinforcement learning: Tutorial, review, and perspectives on open problems,'' \emph{arXiv preprint arXiv:2005.01643}, 2020.

\bibitem{Mnih:2015:DQN}
V.~Mnih, K.~Kavukcuoglu, D.~Silver, A.~A. Rusu, J.~Veness, M.~G. Bellemare, A.~Graves, M.~Riedmiller, A.~K. Fidjeland, G.~Ostrovski, S.~Petersen, C.~Beattie, A.~Sadik, I.~Antonoglou, H.~King, D.~Kumaran, D.~Wierstra, S.~Legg, and D.~Hassabis, ``Human-level control through deep reinforcement learning,'' \emph{Nature}, vol. 518, no. 7540, pp. 529--533, 2015.

\bibitem{Fujimoto2019Off}
S.~Fujimoto, D.~Meger, and D.~Precup, ``Off-policy deep reinforcement learning without exploration,'' in \emph{International conference on machine learning}.\hskip 1em plus 0.5em minus 0.4em\relax PMLR, 2019, pp. 2052--2062.

\bibitem{kumar2020conservative}
A.~Kumar, A.~Zhou, G.~Tucker, and S.~Levine, ``Conservative q-learning for offline reinforcement learning,'' \emph{Advances in Neural Information Processing Systems}, vol.~33, pp. 1179--1191, 2020.

\bibitem{kostrikov2021offline}
I.~Kostrikov, R.~Fergus, J.~Tompson, and O.~Nachum, ``Offline reinforcement learning with fisher divergence critic regularization,'' in \emph{International Conference on Machine Learning}.\hskip 1em plus 0.5em minus 0.4em\relax PMLR, 2021, pp. 5774--5783.

\bibitem{Peng2019Advantage}
X.~B. Peng, A.~Kumar, G.~Zhang, and S.~Levine, ``Advantage-weighted regression: Simple and scalable off-policy reinforcement learning,'' \emph{arXiv preprint arXiv:1910.00177}, 2019.

\bibitem{Chen2020BAIL}
X.~Chen, Z.~Zhou, Z.~Wang, C.~Wang, Y.~Wu, and K.~Ross, ``{BAIL}: Best-action imitation learning for batch deep reinforcement learning,'' \emph{Advances in Neural Information Processing Systems}, vol.~33, pp. 18\,353--18\,363, 2020.

\bibitem{Kidambi2020MoReL}
R.~Kidambi, A.~Rajeswaran, P.~Netrapalli, and T.~Joachims, ``{MoReL}: Model-based offline reinforcement learning,'' \emph{Advances in neural information processing systems}, vol.~33, pp. 21\,810--21\,823, 2020.

\bibitem{Yu2021COMBO}
T.~Yu, A.~Kumar, R.~Rafailov, A.~Rajeswaran, S.~Levine, and C.~Finn, ``{COMBO}: Conservative offline model-based policy optimization,'' \emph{Advances in neural information processing systems}, vol.~34, pp. 28\,954--28\,967, 2021.

\bibitem{Gertz2003Object}
E.~M. Gertz and S.~J. Wright, ``Object-oriented software for quadratic programming,'' \emph{ACM Transactions on Mathematical Software (TOMS)}, vol.~29, no.~1, pp. 58--81, 2003.

\bibitem{Ferreau2014qpOASES}
H.~J. Ferreau, C.~Kirches, A.~Potschka, H.~G. Bock, and M.~Diehl, ``qpoases: A parametric active-set algorithm for quadratic programming,'' \emph{Mathematical Programming Computation}, vol.~6, pp. 327--363, 2014.

\bibitem{Stellato2020Osqp}
B.~Stellato, G.~Banjac, P.~Goulart, A.~Bemporad, and S.~Boyd, ``{OSQP}: An operator splitting solver for quadratic programs,'' \emph{Mathematical Programming Computation}, vol.~12, no.~4, pp. 637--672, 2020.

\bibitem{mansoury2022exposure}
M.~Mansoury, B.~Mobasher, and H.~van Hoof, ``Exposure-aware recommendation using contextual bandits,'' \emph{arXiv preprint arXiv:2209.01665}, 2022.

\bibitem{Pedregosa2011Scikit}
F.~Pedregosa, G.~Varoquaux, A.~Gramfort, V.~Michel, B.~Thirion, O.~Grisel, M.~Blondel, P.~Prettenhofer, R.~Weiss, V.~Dubourg, J.~Vanderplas, A.~Passos, D.~Cournapeau, M.~Brucher, M.~Perrot, and E.~Duchesnay, ``Scikit-learn: Machine learning in {P}ython,'' \emph{Journal of Machine Learning Research}, vol.~12, pp. 2825--2830, 2011.

\bibitem{Wang2019DGL}
M.~Wang, D.~Zheng, Z.~Ye, Q.~Gan, M.~Li, X.~Song, J.~Zhou, C.~Ma, L.~Yu, Y.~Gai, T.~Xiao, T.~He, G.~Karypis, J.~Li, and Z.~Zhang, ``Deep graph library: A graph-centric, highly-performant package for graph neural networks,'' \emph{arXiv preprint arXiv:1909.01315}, 2019.

\bibitem{Koren2009Matrix}
Y.~Koren, R.~Bell, and C.~Volinsky, ``Matrix factorization techniques for recommender systems,'' \emph{Computer}, vol.~42, no.~8, pp. 30--37, 2009.

\bibitem{menendez1997jensen}
M.~Men{\'e}ndez, J.~Pardo, L.~Pardo, and M.~Pardo, ``The jensen-shannon divergence,'' \emph{Journal of the Franklin Institute}, vol. 334, no.~2, pp. 307--318, 1997.

\end{thebibliography}
\begin{IEEEbiography}[{\includegraphics[width=1in,height=1in,clip,keepaspectratio]{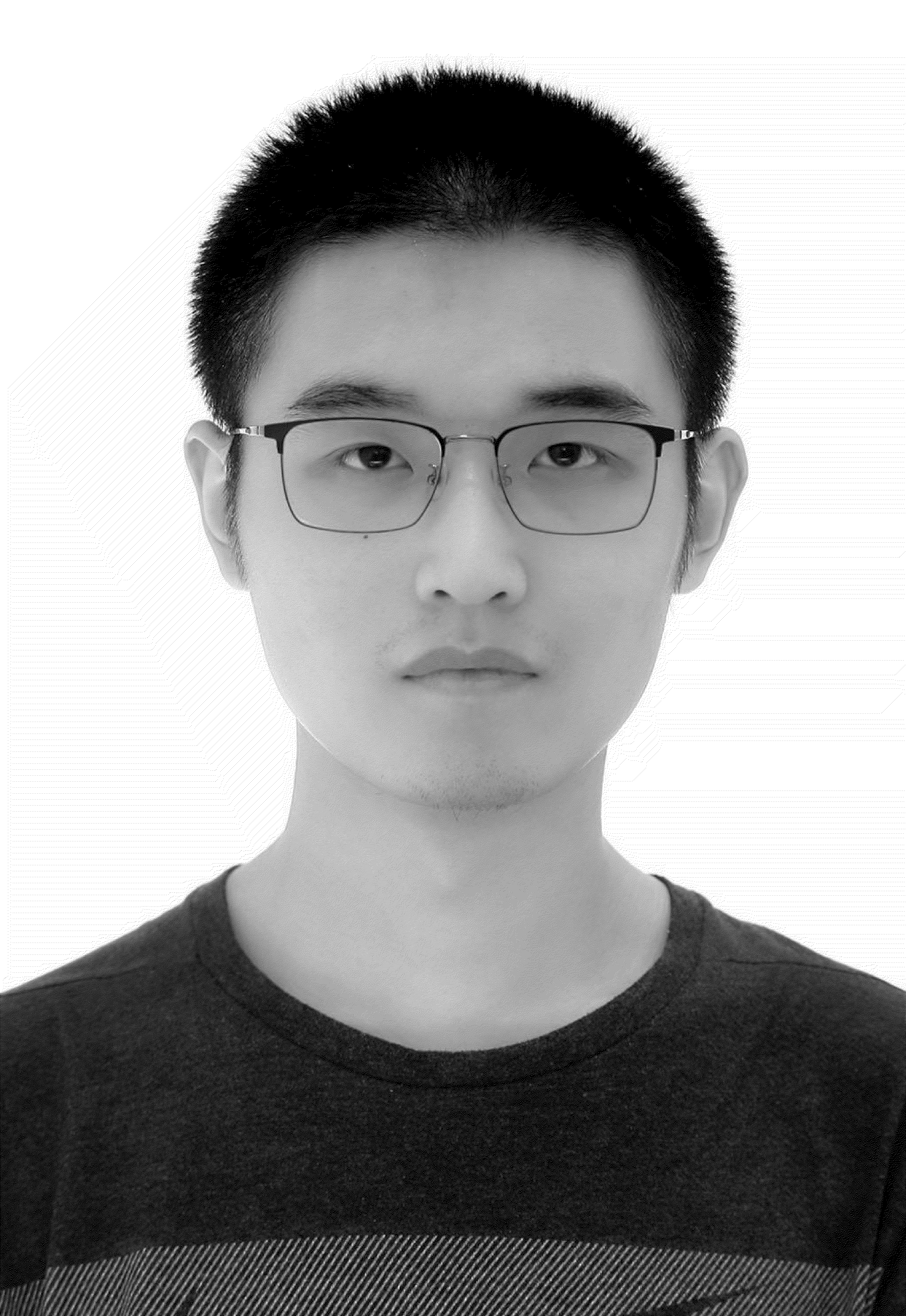}}]
{Yi Xu} is currently pursuing his M.S. degree of Artificial Intelligence at Gaoling School of Artificial Intelligence, Renmin University of China (RUC). His current research interests lie at reinforcement learning, imitation learning and recommender system via Large Language Models' reflections.
\end{IEEEbiography}

\begin{IEEEbiography}[{\includegraphics[width=1in,height=1in,clip,keepaspectratio]{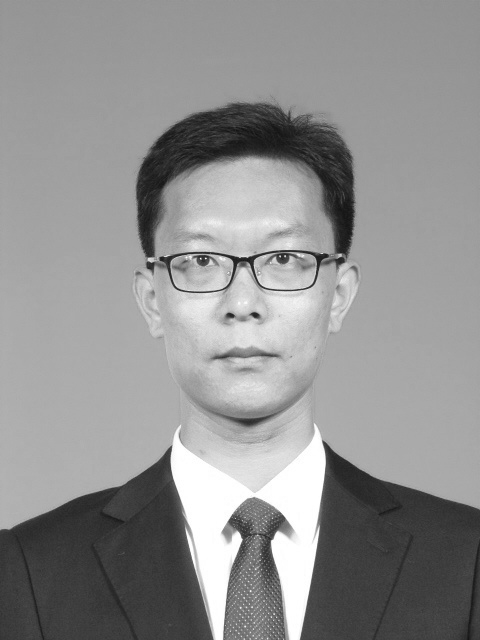}}]
{Weiran Shen} is a tenure-track associate professor at Gaoling School of Artificial Intelligence, Renmin University of China. His research interest includes multi-agent system, game theory, mechanism design, and machine learning. He published over 30 papers in top-tier conferences and journals in these research areas. He was PC, SPC, AC members of multiple international conferences including AAAI, IJCAI, ICML, NeurIPS, WWW, AAMAS.
\end{IEEEbiography}

\begin{IEEEbiography}[{\includegraphics[width=1in,height=1.1in,clip,keepaspectratio]{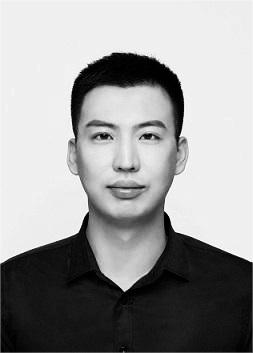}}]
{Xiao Zhang} is a tenure-track associate professor at Gaoling School of Artificial Intelligence, Renmin University of China. His research interests include online learning, trustworthy machine learning, and information retrieval. He has published over 40 papers on top-tier conferences and journals in artificial intelligence, e.g., NeurIPS, ICML, KDD, SIGIR, AAAI, IJCAI, ICDE, WWW, VLDB, etc.
\end{IEEEbiography}

\begin{IEEEbiography}[{\includegraphics[width=1in,height=1in,clip,keepaspectratio]{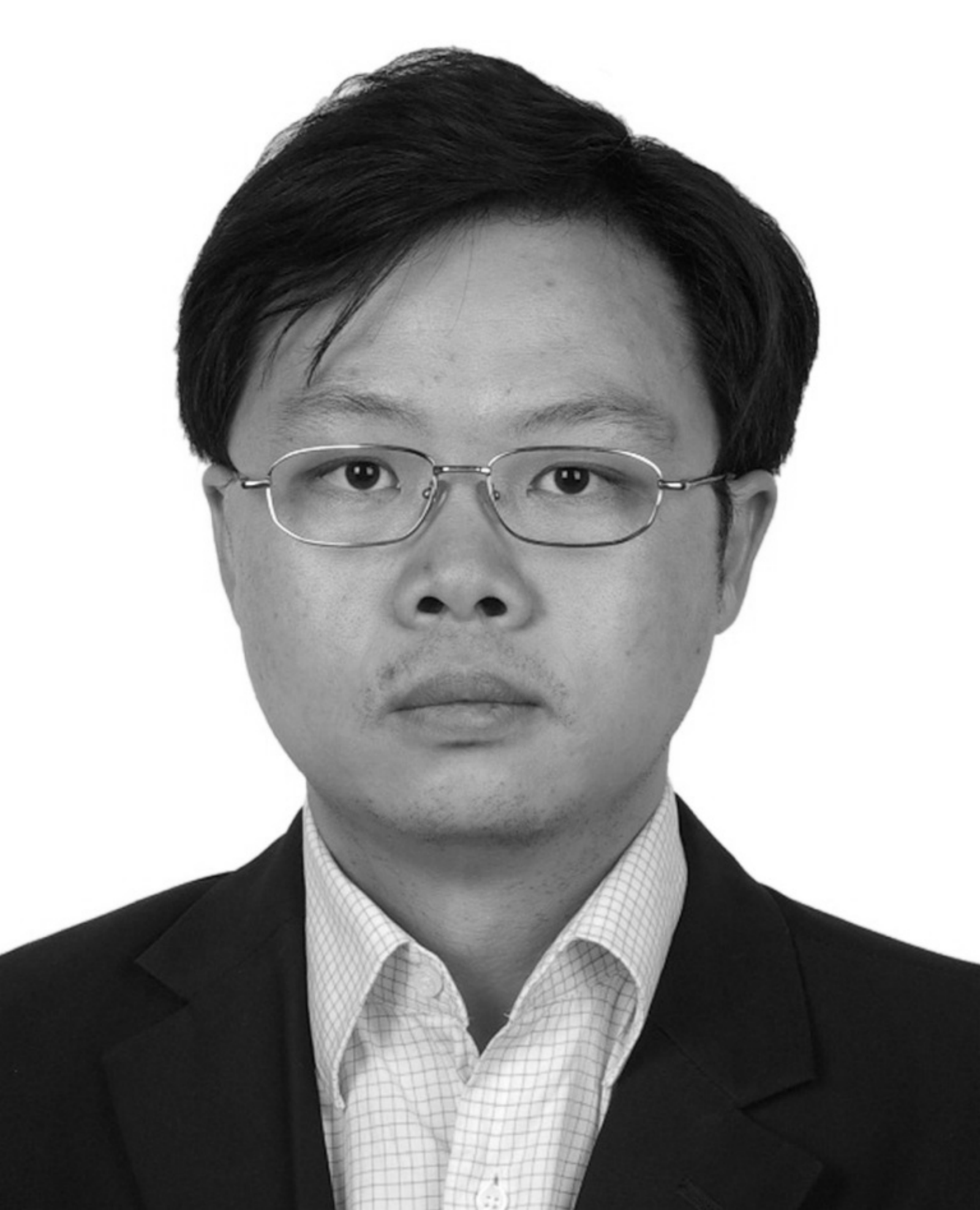}}]{Jun Xu}  is a professor with the Gaoling School of Artiﬁcial Intelligence, Renmin University of China. His research interests focus on learning to rank and semantic matching in web search. He served or is serving as SPC for SIGIR, WWW, and AAAI, editorial board member for Journal of the Association for Information Science and Technology, and associate editor for ACM TIST. He has won the Test of Time Award Honorable Mention in SIGIR (2019), Best Paper Award in AIRS (2010) and Best Paper
\end{IEEEbiography}

\begin{IEEEbiography}
[{\includegraphics[width=1in,height=1in,clip,keepaspectratio]{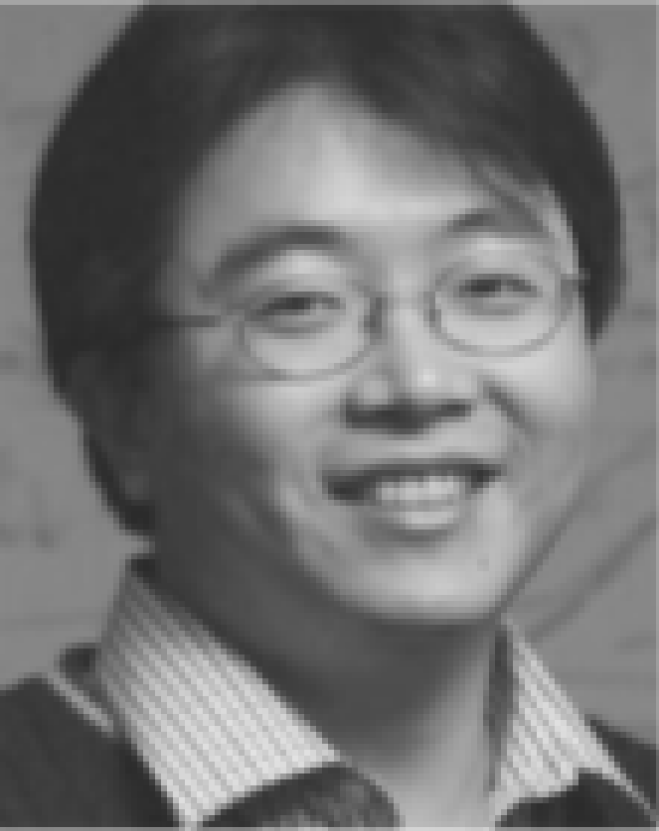}}]{Ji-Rong Wen} is a professor of the Renmin University
of China (RUC). He is also the dean of the School of
Information and executive dean of the Gaoling School
of Artiﬁcial Intelligence with RUC. His main research
interests include information retrieval, data mining,
and machine learning. He was a senior researcher and
group manager of the Web Search and Mining Group
with Microsoft Research Asia (MSRA).
\end{IEEEbiography}

\end{document}